\documentclass[10pt,twocolumn,letterpaper]{article}

\usepackage[pagenumbers]{cvpr}

\usepackage{cuted} %

\usepackage{currfile} %

\usepackage{caption} %

\usepackage{microtype}

\usepackage{xspace}
\usepackage{booktabs}
\usepackage{colortbl}
\definecolor{oursrowgray}{gray}{0.9}
\newcommand{\oursrow}{\rowcolor{oursrowgray}[0pt][2\tabcolsep]}
\makeatletter
\newcommand{\oursend}{\gdef\CT@cell@color{\CT@color{oursrowgray}\@tempdimc\z@\global\let\CT@cell@color\relax}}
\makeatother
\newcommand{\best}[1]{\textbf{#1}}
\usepackage{amsmath,amssymb}
\usepackage{tikz}
\usetikzlibrary{arrows.meta,positioning,calc,shapes.misc}
\definecolor{kolibri}{RGB}{235,104,52}
\definecolor{bufx}{RGB}{42,120,214}
\newcommand{\ours}{Kolibri\xspace}
\newcommand{\fast}{\textit{Fast}\xspace}
\newcommand{\precise}{\textit{Precise}\xspace}
\newcommand{\colorpass}{\textit{color pass}\xspace}

\newif\ifaudit \auditfalse
\newcommand{\pnumdef}[2]{\expandafter\def\csname pnum@#1\endcsname{#2}}
\newcommand{\pnum}[1]{\ifcsname pnum@#1\endcsname\ifaudit\textcolor{green!50!black}{\csname pnum@#1\endcsname}\else\csname pnum@#1\endcsname\fi\else\textbf{\textcolor{red}{[#1]}}\PackageWarning{paper}{Undefined number #1}\fi}
\IfFileExists{results/numbers.tex}{\pnumdef{time.fast.median_ms}{82}
\pnumdef{time.precise.median_ms}{116}
\pnumdef{time.bufferx.median_ms}{1,444}
\pnumdef{time.gedi.median_ms}{24,179}
\pnumdef{speedup.bufferx_over_fast}{18}
\pnumdef{speedup.gedi_over_fast}{296}
\pnumdef{bonn.frames}{597}
\pnumdef{bonn.fast.r50}{26.3}
\pnumdef{bonn.fast.r5}{9.1}
\pnumdef{bonn.fast_color.r50}{47.2}
\pnumdef{bonn.fast_color.r5}{17.3}
\pnumdef{bonn.gedi_tiles.r50}{91.5}
\pnumdef{bonn.gedi_tiles.r5}{42.2}
\pnumdef{bonn.gedi_tiles.min}{11}
\pnumdef{bonn.bufferx_tiles.r50}{74.5}
\pnumdef{bonn.bufferx_tiles.r5}{34.5}
\pnumdef{bonn.bufferx_tiles.min}{1}
\pnumdef{building.rooms}{100}
\pnumdef{building.points_m}{15.0}
\pnumdef{building.frames}{7,986}
\pnumdef{building.fast.r5_one}{81.2}
\pnumdef{building.fast.r5_all}{79.9}
\pnumdef{building.fast.loss}{1.3}
\pnumdef{building.fast.loss.lo}{0.8}
\pnumdef{building.fast.loss.hi}{1.8}
\pnumdef{building.precise.r5_one}{85.1}
\pnumdef{building.precise.r5_all}{83.7}
\pnumdef{building.precise.loss}{1.4}
\pnumdef{building.precise.loss.lo}{0.9}
\pnumdef{building.precise.loss.hi}{1.9}
\pnumdef{building.fast.right_room}{98.0}
\pnumdef{building.one_candidate.r5_all}{46.2}
\pnumdef{building.retrieval.top1}{28.6}
\pnumdef{building.retrieval.top3}{44.7}
\pnumdef{building.rr1.r5}{23.9}
\pnumdef{building.fast_minus_rr1}{56.0}
\pnumdef{building.fast_minus_rr1.lo}{53.3}
\pnumdef{building.fast_minus_rr1.hi}{58.6}
\pnumdef{building.timing_frames}{288}
\pnumdef{building.fast.ms_one}{65}
\pnumdef{building.fast.ms_all}{168}
\pnumdef{building.precise.ms_all}{230}
\pnumdef{building.peak_gb}{2.15}
\pnumdef{building.bufferx.s_one}{1.3}
\pnumdef{building.bufferx.s_all}{46}
\pnumdef{building.bufferx.zero_from}{25}
\pnumdef{building.color_oom_rooms}{10}
\pnumdef{cam.frames}{871}
\pnumdef{cam.native_hfov}{58}
\pnumdef{cam.fov45.kept}{59}
\pnumdef{cam.fov30.kept}{25}
\pnumdef{cam.fast.clean.r25}{86.9}
\pnumdef{cam.fast.focal110.r25}{30.3}
\pnumdef{cam.gedi.focal110.r25}{29.5}
\pnumdef{cam.bx.focal110.r25}{30.3}
\pnumdef{cam.fast.focal.min_r50}{90.3}
\pnumdef{cam.precise_below_fast.focal}{4}
\pnumdef{cam.calib.comparisons}{16}
\pnumdef{cam.calib.equivalent}{16}
\pnumdef{cam.fast.dist.max_drop}{8.0}
\pnumdef{cam.fast.color.max_change}{0.5}
\pnumdef{cam.fov30.fast_minus_gedi}{6.2}
\pnumdef{cam.fov30.fast_minus_gedi.lo}{+2.7}
\pnumdef{cam.fov30.fast_minus_gedi.hi}{+10.1}
\pnumdef{cam.fov30.fast_minus_bx}{9.6}
\pnumdef{cam.fov30.fast_minus_bx.lo}{+5.6}
\pnumdef{cam.fov30.fast_minus_bx.hi}{+13.8}
\pnumdef{cam.fov45.deg}{45}
\pnumdef{cam.fov30.deg}{30}
\pnumdef{cam.focal.small}{5}
\pnumdef{cam.focal.large}{10}
\pnumdef{cam.focal.symmetric}{True}
\pnumdef{cam.color.small}{5}
\pnumdef{cam.color.large}{10}
\pnumdef{cam.margin}{3}
\pnumdef{campaign_a.rooms}{96}
\pnumdef{campaign_a.rooms_checked}{100}
\pnumdef{campaign_a.frames}{7,986}
\pnumdef{campaign_a.rescan.frames}{5,133}
\pnumdef{campaign_a.rescan.rooms}{41}
\pnumdef{campaign_a.rescan.scans}{60}
\pnumdef{campaign_a.fast.r5}{81.4}
\pnumdef{campaign_a.precise.r5}{85.3}
\pnumdef{campaign_a.fast_color.r5}{81.6}
\pnumdef{campaign_a.precise_color.r5}{85.5}
\pnumdef{campaign_a.gedi.r5}{80.3}
\pnumdef{campaign_a.colorpcr.r5}{80.3}
\pnumdef{campaign_a.geotrrgb.r5}{77.8}
\pnumdef{campaign_a.roitr.r5}{75.2}
\pnumdef{campaign_a.bufferx.r5}{69.4}
\pnumdef{campaign_a.lite.r5}{70.4}
\pnumdef{campaign_a.fast_minus_bufferx}{11.9}
\pnumdef{campaign_a.fast_minus_gedi}{+1.1}
\pnumdef{campaign_a.fast_minus_gedi.lo}{+0.6}
\pnumdef{campaign_a.fast_minus_gedi.hi}{+1.6}
\pnumdef{campaign_a.precise_minus_fast}{4.0}
\pnumdef{campaign_a.precise_minus_fast.lo}{3.3}
\pnumdef{campaign_a.precise_minus_fast.hi}{4.6}
\pnumdef{campaign_a.bufferx_zero_rooms}{6}
\pnumdef{campaign_a.matched.cells}{240}
\pnumdef{campaign_a.matched.trail}{0}
\pnumdef{campaign_a.rescan.fast.r5}{65.1}
\pnumdef{campaign_a.rescan.gedi.r5}{65.5}
\pnumdef{campaign_a.rescan.fast_minus_bufferx}{3.6}
\pnumdef{campaign_a.rescan.cost_min}{16}
\pnumdef{campaign_a.rescan.cost_max}{25}
\pnumdef{campaign_a.rescan.changed_n}{243}
\pnumdef{campaign_a.rescan.changed.fast_r05}{57.9}
\pnumdef{campaign_a.rescan.changed.gedi_r05}{59.7}
\pnumdef{campaign_a.rescan.changed.colorpcr_r05}{64.6}
\pnumdef{campaign_a.rescan.changed.bufferx_r05}{47.7}
\pnumdef{cells.fine_cm}{5}
\pnumdef{cells.coarse_cm}{20}
\pnumdef{cells.c05.loss_heldout}{2.3}
\pnumdef{cells.c20.loss_heldout}{1.9}
\pnumdef{cells.c05.oom_pct}{40}
\pnumdef{cells.c10.ms}{82}
\pnumdef{cells.c05.ms}{249}
\pnumdef{cells.c20.ms}{90}
\pnumdef{cells.c05.time_ratio}{3}
\pnumdef{depthcp.comparisons}{144}
\pnumdef{depthcp.fast.scannet_r5_gain}{1.1}
\pnumdef{depthcp.fast.scannet_r5_gain.lo}{0.4}
\pnumdef{depthcp.fast.scannet_r5_gain.hi}{2.0}
\pnumdef{diag.rooms}{29}
\pnumdef{diag.gray.rho}{+0.63}
\pnumdef{diag.gray.lo}{+0.38}
\pnumdef{diag.gray.hi}{+0.79}
\pnumdef{diag.gain.rho}{-0.27}
\pnumdef{diag.gain.lo}{-0.66}
\pnumdef{diag.gain.hi}{+0.23}
\pnumdef{diag.within.scannet_gray}{+0.66}
\pnumdef{diag.within.sevenscenes_gray}{+0.75}
\pnumdef{method.params}{55,712}
\pnumdef{method.cell_cm}{10}
\pnumdef{method.k}{16}
\pnumdef{method.desc_dim}{32}
\pnumdef{method.lookalike_m}{1.06}
\pnumdef{method.lookalike_deg}{177}
\pnumdef{method.err_cm}{1.7}
\pnumdef{method.fit_true}{0.95}
\pnumdef{method.fit_look}{0.79}
\pnumdef{method.verify_cm}{3}
\pnumdef{method.candidates}{6}
\pnumdef{method.fit_points}{1,000}
\pnumdef{method.timing_frames}{200}
\pnumdef{method.network_ms}{1.9}
\pnumdef{method.precise_plus_ms}{35}
\pnumdef{method.colorpass_ratio}{2.3}
\pnumdef{robust.depth.levels}{12}
\pnumdef{robust.depth.trails}{0}
\pnumdef{robust.depth.bufferx_max}{8.6}
\pnumdef{robust.depth.gedi_max}{34.3}
\pnumdef{robust.color.levels}{15}
\pnumdef{robust.color.pass_gain}{10.7}
\pnumdef{robust.color.colorpcr_mean}{26.9}
\pnumdef{robust.color.geotr_mean}{9.2}
\pnumdef{robust.color.geotr_clean}{3.9}
\pnumdef{robust.color.geotr_gray}{47.2}
\pnumdef{robust.color.geotr_cnoise}{39.9}
\pnumdef{rooms.scannet.fast.recall5}{74.3}
\pnumdef{rooms.scannet.precise.recall5}{78.1}
\pnumdef{rooms.scannet.oracle.recall5}{74.7}
\pnumdef{rooms.scannet.rr3.recall5}{60.3}
\pnumdef{rooms.scannet.rr1.recall5}{52.2}
\pnumdef{rooms.seven.fast.recall5}{90.4}
\pnumdef{rooms.seven.oracle.recall5}{92.2}
\pnumdef{rooms.seven.rr3.recall5}{85.3}
\pnumdef{rooms.scannet.fast.right_room}{97.9}
\pnumdef{rooms.scannet.retrieval.top3}{80.8}
\pnumdef{rooms.scannet.retrieval.top1}{67.9}
\pnumdef{rooms.scannet.bufferx_whole.recall5}{9.9}
\pnumdef{rooms.scannet.fast_minus_rr3}{14.0}
\pnumdef{rooms.scannet.fast_minus_rr3.lo}{+8.4}
\pnumdef{rooms.scannet.fast_minus_rr3.hi}{+19.9}
\pnumdef{rooms.scannet.fast_minus_oracle}{-0.5}
\pnumdef{rooms.seven.fast_minus_rr3}{5.1}
\pnumdef{rooms.scannet.fast_minus_oracle.abs}{0.5}
\pnumdef{rooms.scannet.fast.ms}{99}
\pnumdef{rooms.scannet.rr3.s}{4.6}
\pnumdef{rooms.scannet.rr3_over_fast}{47}
\pnumdef{rooms.mapsize.peak_gb}{2.1}
\pnumdef{rooms.mapsize.bufferx.s_one}{1.4}
\pnumdef{rooms.mapsize.bufferx.s_all}{5.5}
\pnumdef{rooms.scannet.oracle2.recall5}{77.4}
\pnumdef{rooms.scannet.precise_minus_oracle2}{+0.8}
\pnumdef{small.basin.e50_cm}{7.7}
\pnumdef{small.basin.both_inside_pct}{62}
\pnumdef{small.basin.same_outcome_pct}{99}
\pnumdef{small.size.params_min}{4,736}
\pnumdef{small.size.params_max}{915,584}
\pnumdef{small.size.params_ours}{55,712}
\pnumdef{small.size.max_abs_diff}{1.0}
\pnumdef{small.size.ms_min}{81}
\pnumdef{small.size.ms_max}{94}
\pnumdef{small.size.l0_recall}{63.8}
\pnumdef{small.maps.min}{5}
\pnumdef{small.maps.min_recall}{76.4}
\pnumdef{small.maps.second}{22}
\pnumdef{small.maps.second_recall}{78.3}
\pnumdef{small.maps.max}{170}
\pnumdef{small.maps.max_recall}{78.6}
\pnumdef{teaser.fast.recall5}{75.7}
\pnumdef{teaser.reach.ratio}{15}
\pnumdef{teaser.price.fast_gray}{30.6}
\pnumdef{teaser.price.geotr_gray}{73.7}
\pnumdef{teaser.price.gray_gap}{43}
\pnumdef{final.rooms}{10}
\pnumdef{final.real.n}{679}
\pnumdef{final.rendered.n}{2,000}
\pnumdef{final.real.precise.r5}{73.1}
\pnumdef{final.real.precise_vs_gedi2.base_r5}{72.6}
\pnumdef{final.real.precise_minus_gedi2}{+0.5}
\pnumdef{final.real.precise_minus_gedi2.lo90}{-1.7}
\pnumdef{final.real.precise_minus_gedi2.hi90}{+2.4}
\pnumdef{final.real.precise_minus_gedi2.verdict}{equivalent}
\pnumdef{final.real.fast.r5}{68.7}
\pnumdef{final.real.fast_vs_gedi.base_r5}{68.6}
\pnumdef{final.real.fast_minus_gedi}{+0.0}
\pnumdef{final.real.fast_minus_gedi.lo90}{-2.1}
\pnumdef{final.real.fast_minus_gedi.hi90}{+2.1}
\pnumdef{final.real.fast_minus_gedi.verdict}{equivalent}
\pnumdef{final.real.precise_color.r5}{73.4}
\pnumdef{final.real.precise_color_vs_gedi2.base_r5}{72.6}
\pnumdef{final.real.precise_color_minus_gedi2}{+0.8}
\pnumdef{final.real.precise_color_minus_gedi2.lo90}{-1.5}
\pnumdef{final.real.precise_color_minus_gedi2.hi90}{+2.7}
\pnumdef{final.real.precise_color_minus_gedi2.verdict}{equivalent}
\pnumdef{final.rendered.precise.r50}{95.0}
\pnumdef{final.rendered.colorpcr.r50}{94.3}
\pnumdef{final.rendered.precise_minus_colorpcr.verdict}{equivalent}
\pnumdef{final.rendered.gedi.r50}{91.0}
\pnumdef{final.rendered.precise_minus_gedi.verdict}{lead}
\pnumdef{final.rendered.bufferx.r50}{90.8}
\pnumdef{final.rendered.precise_minus_bufferx.verdict}{lead}
\pnumdef{hosp.points_m}{2.1}
\pnumdef{hosp.extent_x}{49}
\pnumdef{hosp.extent_y}{42}
\pnumdef{hosp.frames}{440}
\pnumdef{hosp.walks}{30}
\pnumdef{hosp.sub_frames}{30}
\pnumdef{hosp.tiles}{1,090}
\pnumdef{hosp.tile_m}{3}
\pnumdef{hosp.tile_stride_m}{1.5}
\pnumdef{hosp.cells}{350,087}
\pnumdef{hosp.fast.r5}{86.5}
\pnumdef{hosp.fast.r50}{90.6}
\pnumdef{hosp.fast.r5_sub}{84.4}
\pnumdef{hosp.precise.r5}{86.4}
\pnumdef{hosp.precise.r50}{91.1}
\pnumdef{hosp.precise.r5_sub}{86.7}
\pnumdef{hosp.fast.ms}{384}
\pnumdef{hosp.precise.ms}{621}
\pnumdef{hosp.fast.peak_gb}{2.3}
\pnumdef{hosp.gedi_whole.r5}{26.7}
\pnumdef{hosp.gedi_whole.r5_2}{30.0}
\pnumdef{hosp.gedi_whole.min}{9}
\pnumdef{hosp.gedi_tiles.r5}{76.7}
\pnumdef{hosp.gedi_tiles.r5_2}{80.0}
\pnumdef{hosp.gedi_tiles.min}{91}
\pnumdef{hosp.bufferx_tiles.r5}{60.0}
\pnumdef{hosp.bufferx_tiles.r5_2}{63.3}
\pnumdef{hosp.bufferx_tiles.min}{16}
\pnumdef{hosp.gedi_sparse.r5}{43.3}
\pnumdef{hosp.gedi_sparse.r5_2}{46.7}
\pnumdef{hosp.gedi_sparse.min}{2}
\pnumdef{hosp.gedi_sparse.keypoints}{350,000}
\pnumdef{hosp.gedi_sparse.rd_cm}{15}
\pnumdef{hosp.gedi_sparse.n_rd}{5}
\pnumdef{hosp.fast_minus_gedi_tiles}{+7.8}
\pnumdef{hosp.fast_minus_gedi_tiles.lo}{-7.8}
\pnumdef{hosp.fast_minus_gedi_tiles.hi}{+23.4}
\pnumdef{hosp.fast_minus_gedi_whole}{+57.8}
\pnumdef{hosp.fast_minus_gedi_whole.lo}{+41.1}
\pnumdef{hosp.fast_minus_gedi_whole.hi}{+74.4}
\pnumdef{hosp.gedi_tiles_over_fast}{14,275}
\pnumdef{hosp.gedi_sparse_over_fast}{315}
\pnumdef{method.consensus_den}{10}
\pnumdef{method.consensus_min}{8}
\pnumdef{method.distinct_m}{0.5}
\pnumdef{method.distinct_deg}{10}
\pnumdef{method.icp_radius_end_cm}{5}
\pnumdef{method.min_ratio_pct}{50}
\pnumdef{method.positive_m}{0.5}
\pnumdef{method.piece_cells}{16,384}
\pnumdef{train.depth_noise}{0.005}
\pnumdef{train.noise_cm}{1}
\pnumdef{train.scale_pct}{15}
\pnumdef{train.color_jitter}{0.3}
\pnumdef{train.crop_pct}{30}
\pnumdef{prose.net.width}{128}
\pnumdef{prose.net.layers}{1}
\pnumdef{prose.net.temperature}{0.1}
\pnumdef{prose.train.lr}{0.001}
\pnumdef{prose.train.batch_views}{8}
\pnumdef{prose.train.steps}{18,600}
\pnumdef{prose.train.minutes_max}{53}
\pnumdef{prose.train.scannet_rooms}{165}
\pnumdef{prose.train.maps}{170}
\pnumdef{prose.train.building_maps}{5}
\pnumdef{prose.cells.frame_mean}{949}
\pnumdef{prose.cells.map_median}{6,660}
\pnumdef{prose.est.topk}{5}
\pnumdef{prose.est.rho_cm}{25}
\pnumdef{prose.est.seeds}{256}
\pnumdef{prose.est.first_round}{4}
\pnumdef{prose.est.rounds}{2}
\pnumdef{prose.est.peel_m}{0.5}
\pnumdef{prose.est.icp_iters}{10}
\pnumdef{prose.est.icp_points}{1,000}
\pnumdef{prose.est.icp_radius_cm}{20}
\pnumdef{prose.second.radius_cm}{10}
\pnumdef{prose.second.scale_cm}{5}
\pnumdef{prose.time.describe_ms}{3.8}
\pnumdef{prose.time.map_ms}{25}
\pnumdef{prose.time.coarse_ms}{22.5}
\pnumdef{prose.time.bufferx_s}{1.4}
\pnumdef{prose.time.gedi_s}{24}
\pnumdef{prose.time.fast_p95}{251}
\pnumdef{prose.time.bufferx_p95}{1,948}
\pnumdef{prose.frames.scannet}{871}
\pnumdef{prose.frames.sevenscenes}{1,700}
\pnumdef{prose.frames.twelvescenes}{575}
\pnumdef{prose.frames.xsession}{683}
\pnumdef{prose.t1.strong.cells}{27}
\pnumdef{prose.t1.strong_precise.cells}{27}
\pnumdef{prose.t1.strong.equivalent}{14}
\pnumdef{prose.t1.strong_precise.equivalent}{11}
\pnumdef{prose.t1.strong.lead}{8}
\pnumdef{prose.t1.strong_precise.lead}{11}
\pnumdef{prose.t1.strong.inconclusive_pos}{5}
\pnumdef{prose.t1.strong_precise.inconclusive_pos}{5}
\pnumdef{prose.t1.strong.inconclusive_neg}{0}
\pnumdef{prose.t1.strong_precise.inconclusive_neg}{0}
\pnumdef{prose.t1.strong.trail}{0}
\pnumdef{prose.t1.strong_precise.trail}{0}
\pnumdef{prose.t1.all.cells}{96}
\pnumdef{prose.t1.all.trail}{0}
\pnumdef{prose.t1.native.trail}{0}
\pnumdef{prose.t1.twelve.bufferx_lead}{6.1}
\pnumdef{prose.t1.twelve.lite_lead}{6.3}
\pnumdef{prose.t1.fast.sevenscenes_r5}{93.0}
\pnumdef{prose.t1.bufferx_native.sevenscenes_r5}{72.7}
\pnumdef{prose.t1.bufferx_icp.sevenscenes_r5}{92.2}
\pnumdef{prose.front.settings}{29}
\pnumdef{prose.front.match_ms}{2,543}
\pnumdef{prose.front.match_ratio}{31}
\pnumdef{prose.front.cached_ms}{414}
\pnumdef{prose.front.cached_ratio}{5.1}
\pnumdef{prose.front.cached_twelve_gap}{7.3}
\pnumdef{prose.front.fastest_ms}{136}
\pnumdef{prose.front.fastest_loss_min}{10}
\pnumdef{prose.front.fastest_loss_max}{28}
\pnumdef{prose.basin.end_mm}{0.03}
\pnumdef{prose.basin.outside_ours}{51}
\pnumdef{prose.basin.outside_bufferx}{21}
\pnumdef{prose.swap.fpfh}{30.2}
\pnumdef{prose.swap.fpfh_color}{61.8}
\pnumdef{prose.swap.untrained}{62.6}
\pnumdef{prose.swap.ours}{78.5}
\pnumdef{prose.swap.gedi}{79.4}
\pnumdef{prose.swap.training_gain}{16}
\pnumdef{prose.swap.gedi_coarse}{64.1}
\pnumdef{prose.swap.ours_coarse}{46.7}
\pnumdef{prose.room.n}{29}
\pnumdef{prose.room.trail_bufferx}{0}
\pnumdef{prose.room.trail_gedi}{1}
\pnumdef{prose.xs.second_gain_min}{2.0}
\pnumdef{prose.xs.second_gain_max}{2.9}
\pnumdef{prose.xs.held_n}{640}
\pnumdef{prose.xs.changed_n}{43}
\pnumdef{prose.xs.held.fast_r5}{77.3}
\pnumdef{prose.xs.changed.fast_r05}{54.3}
\pnumdef{prose.xs.held.bufferx_r5}{76.9}
\pnumdef{prose.xs.changed.bufferx_r05}{83.7}
\pnumdef{prose.xs.held.gedi_r5}{77.0}
\pnumdef{prose.xs.changed.gedi_r05}{88.4}
\pnumdef{prose.rooms.pitch_m}{20}
\pnumdef{prose.rooms.time_ratio}{1.19}
\pnumdef{prose.rooms.retrieval_ms}{11}
\pnumdef{prose.rooms.retrieval_views}{64}
\pnumdef{prose.rooms.retrieval_vote}{5}
\pnumdef{tables.seed_sd_max}{0.6}
\pnumdef{sets.scannet10.rooms}{10}
\pnumdef{sets.sevenscenes.rooms}{7}
\pnumdef{sets.twelvescenes.rooms}{12}
\pnumdef{sets.xsession.rooms}{4}
\pnumdef{sets.xsession.scans}{7}
\pnumdef{bx.cap_m}{5}
\pnumdef{bx.k1.r_big}{0.75}
\pnumdef{bx.k2.r_big}{1.08}
\pnumdef{bx.k5.r_big}{2.50}
\pnumdef{bx.cap_from_rooms}{10}
\pnumdef{bx.scale.frames}{871}
\pnumdef{bx.scale.official.r5}{74.7}
\pnumdef{bx.scale.official.r50}{92.9}
\pnumdef{bx.scale.official.r_big}{0.84}
\pnumdef{bx.scale.frame.r5}{72.2}
\pnumdef{bx.scale.frame.r50}{90.0}
\pnumdef{bx.scale.frame.r_big}{0.33}
\pnumdef{hosp.gedi_grid.runs}{11}
\pnumdef{hosp.gedi_grid.best_r5}{43.3}
\pnumdef{hosp.export_points_m}{1.92}
\pnumdef{gedi.cal_cm}{2}
\pnumdef{bonn.gedi_whole.r50}{77.1}
\pnumdef{bonn.gedi_whole.min}{2.8}
\pnumdef{hosp.max_steps}{20}
\pnumdef{hosp.step_m}{0.5}
\pnumdef{hosp.cam_h_lo_m}{1.5}
\pnumdef{hosp.cam_h_hi_m}{1.7}
\pnumdef{hosp.pitch_max_deg}{20}
\pnumdef{hosp.splat_half_cm}{2}
\pnumdef{hosp.min_cover_pct}{30}
\pnumdef{hosp.min_median_m}{0.8}
\pnumdef{hosp.crop.frames}{30}
\pnumdef{hosp.crop.cube3.r5}{50.0}
\pnumdef{hosp.crop.cube3.inside}{75}
\pnumdef{hosp.crop.sq4.inside}{39}
\pnumdef{hosp.crop.valid.cube3}{100}
\pnumdef{hosp.crop.valid.sq32}{40}
\pnumdef{hosp.crop.sq32.r5}{13.3}
\pnumdef{tab_ablation.precise.no_message_passing_loss}{13.4}
\pnumdef{tab_ablation.precise.no_color_loss}{26.2}
\pnumdef{tab_ablation.precise.color_only_loss}{3.4}
\pnumdef{tab_ablation.fast.gedi_diff}{+0.9}
\pnumdef{tab_ablation.gedi_param_ratio}{56}
\pnumdef{tab_ablation.untrained_vs_fpfh_color_max}{2.4}
\pnumdef{tab_main.margin}{3}
\pnumdef{tab_main.timing_frames}{200}
\pnumdef{tab_main.same_session.comparisons}{48}
\pnumdef{tab_main.same_session.excluded_losses}{48}
\pnumdef{tab_main.fast.peak_gb}{1.44}
\pnumdef{tab_main.bufferx.peak_gb}{6.21}
\pnumdef{tab_main.gedi.peak_gb}{5.27}
\pnumdef{tab_main.fast.twelvescenes_r5}{97.9}
\pnumdef{tab_main.fast.scannet_r5}{75.7}
\pnumdef{tab_main.native_gedi.scannet_r5}{78.8}
\pnumdef{tab_main.native_geotrrgb.twelvescenes_r5}{99.1}
\pnumdef{tab_main.native_gedi.twelvescenes_r5}{98.8}
\pnumdef{tab_main.xsession.r05_max_gap}{2.2}
\pnumdef{tab_main.threedmatch.twelvescenes_rooms}{8}
\pnumdef{tab_main.twelvescenes.rooms}{12}
\pnumdef{ref.dsacstar_rgb.sevenscenes_r5}{84.0}
\pnumdef{ref.dsacstar_rgb.twelvescenes_r5}{99.2}
\pnumdef{ref.dsacstar_rgbd.sevenscenes_r5}{94.7}
\pnumdef{ref.ace.published.sevenscenes_r5}{80.8}
\pnumdef{ref.ace.published.twelvescenes_r5}{99.6}
\pnumdef{ref.ace.run.sevenscenes_full_split_r5}{80.7}
\pnumdef{ref.ace.run.sevenscenes_r5}{76.8}
\pnumdef{ref.ace.run.twelvescenes_r5}{99.5}
\pnumdef{ref.ace.run.mapping_minutes}{4}
\pnumdef{ref.ace.run.twelvescenes_full_split_r5}{99.8}
\pnumdef{ref.ace.run.query_ms}{29}
\pnumdef{ref.ace.run.threads}{8}
\pnumdef{tab_points.precise_extra_ms}{35}
\pnumdef{tab_points.fast_color.median_ms}{188}
\pnumdef{tab_points.precise_color.median_ms}{278}
\pnumdef{tab_points.fast_color.time_ratio}{2.3}
\pnumdef{tab_points.precise_color.time_ratio}{2.4}
\pnumdef{tab_points.precise.xsession_gain}{2.6}
\pnumdef{tab_points.precise.xsession_gain_lo}{0.9}
\pnumdef{tab_points.precise.xsession_gain_hi}{4.5}
\pnumdef{tab_points.fast_color.color_gain_scannet}{10.7}
\pnumdef{tab_points.fast_color.color_gain_sevenscenes}{11.8}
\pnumdef{tab_points.fastest_learned_native_ms}{728}
\pnumdef{tab_points.slowest_point_speedup}{2.6}
\pnumdef{zs.fast.scannet_r5}{75.7}
\pnumdef{zs.mast3r.scannet_r5}{44.7}
\pnumdef{zs.mast3r.scannet_valid}{57.2}
\pnumdef{zs.reloc3r.scannet_r5}{11.4}
\pnumdef{zs.mast3r.s}{23}
\pnumdef{zs.comparisons}{90}
\pnumdef{zs.lead_min}{5.2}
\pnumdef{zs.lead_max}{92.6}
}{}

\DeclareUnicodeCharacter{25E6}{\ensuremath{{}^\circ}}
\providecommand{\tabpointsdash}{}%

\definecolor{cvprblue}{rgb}{0.21,0.49,0.74}
\usepackage[pagebackref,breaklinks,colorlinks,allcolors=cvprblue]{hyperref}

\def\confName{CVPR}
\def\confYear{2027}

\title{Kolibri: Color Finds the Place, Refinement Sets the Pose for RGB-D Localization in Colored Point Clouds}

\author{Leonel Aguilar\\
\small Chair of Cognitive Science, ETH Z\"urich\\
{\tt\small aleonel@ethz.ch}
\and
Bartu Tuncay\\
\small Chair of Cognitive Science, ETH Z\"urich\\
{\tt\small btuncay@ethz.ch}
\and
Christoph H\"olscher\\
\small Chair of Cognitive Science, ETH Z\"urich\\
{\tt\small choelsch@ethz.ch}
}

\begin{document}
\maketitle
\begin{strip}
\centering
\includegraphics[width=0.95\linewidth]{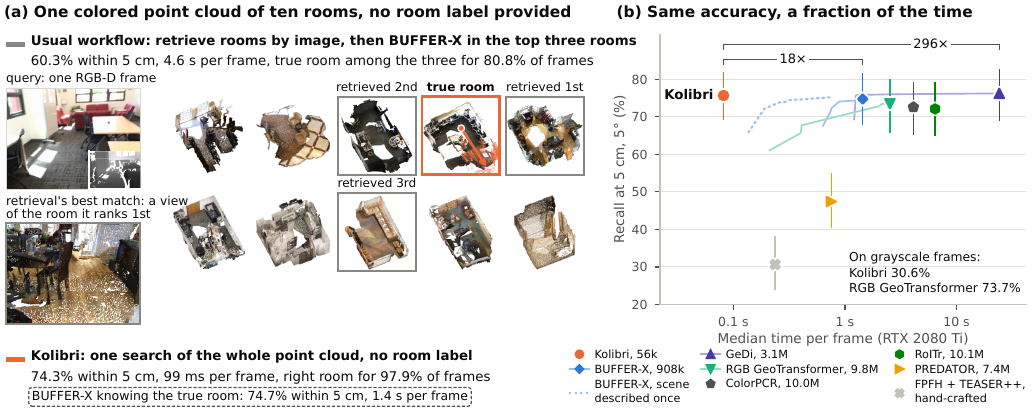}
\captionsetup{hypcap=false}
\captionof{figure}{\textbf{\ours searches a colored point cloud without a room label and matches the accuracy of BUFFER-X given the true room.}
(a)~In the ten-room ScanNet-10 point cloud, image retrieval favors similar-looking views from other rooms (gray).
Searching the full point cloud lets \ours localize this frame in its true room (orange).
(b)~With the room given, recall at 5\,cm and 5\textdegree{} is plotted against median time per frame on an RTX 2080 Ti.
Methods use their released settings and our refinement, with \ours in its standard \fast setting.
Vertical bars are 95\% intervals that resample whole walks of consecutive frames.
Thin lines in a method's color show its other settings (BUFFER-X, GeDi and RGB GeoTransformer).}
\captionsetup{hypcap=true}
\label{fig:teaser}%
\end{strip}

\begin{abstract}
An agent or robot in a scanned building must locate its camera in rooms that no model saw during training.
Point cloud registration methods can align each RGB-D frame with the building's colored point cloud, accurately but at a high cost.
We observe that geometric refinement brings nearby initial poses to the same final pose, so the learned component only needs to find the place.
\ours does this with a graph network of \pnum{method.params} parameters that describes \pnum{method.cell_cm}\,cm cubes of points, called cells, by their color, local shape and neighbors.
With the same refinement, it matches GeDi on \pnum{campaign_a.rooms} unseen ScanNet rooms (\pnum{campaign_a.fast.r5}\% against \pnum{campaign_a.gedi.r5}\% within 5\,cm and 5\textdegree) and on a held-out test of \pnum{final.rooms} further rooms, in \pnum{time.fast.median_ms}\,ms per frame on an RTX 2080 Ti.
The saved time allows for search without a room label.
In one point cloud of \pnum{building.rooms} rooms, \ours loses \pnum{building.fast.loss} points against knowing the room, while retrieving the room first loses \pnum{building.fast_minus_rr1}.
In a hospital emergency department scene from a real scan, it localizes \pnum{hosp.fast.r5}\% of the frames rendered from the scan in \pnum{hosp.fast.ms}\,ms each, where GeDi needs \pnum{hosp.gedi_tiles.min} minutes per frame on cubes of the scan for a similar recall.
The method needs colors that agree between the frame and the point cloud.
On gray frames and across cameras with different color responses, point cloud registration methods keep the lead.
\end{abstract}

\section{Introduction}
\label{sec:intro}

We scanned the emergency department of a university hospital on foot with a handheld laser scanner (BLK2GO) and obtained a colored point cloud of it.
Robots and simulated pedestrian agents need to locate their cameras anywhere in it, without a room label or an initial position.
Each view forms a RGB-D frame, with paired color with depth, so point cloud registration methods can align it with the point cloud.
The difficulty is cost.
On views rendered from the scan with simulated depth noise, GeDi~\cite{poiesi2022learning} needs \pnum{hosp.gedi_tiles.min} minutes per frame to search overlapping cubes of the scan, while \ours, the method we present here, reaches a similar recall on the whole scan in \pnum{hosp.fast.ms}\,ms.
The rendered colors agree with the point cloud by construction, and the baseline comparison uses \pnum{hosp.sub_frames} frames, one per walk (\cref{sec:hospital}).
The hospital scan motivated the method but remained untouched while we developed it on public benchmarks.

A descriptor summarizes the shape or color around a point as a vector, so that the same surface seen twice can be matched.
Point cloud registration methods spend time computing these descriptors and searching for consistent matches.
Even in a single room, BUFFER-X~\cite{seo2025bufferx} takes a median \pnum{prose.time.bufferx_s}\,s per frame and GeDi \pnum{prose.time.gedi_s}\,s on an RTX 2080 Ti when assigned to align overlapping fragments.
At building scale, searching separate rooms or cubes repeats that work.
Image retrieval can narrow the search, but choosing the wrong room loses the frame before pose estimation begins.
Among \pnum{building.rooms} ScanNet rooms, retrieval ranks the true room first for only \pnum{building.retrieval.top1}\% of frames.

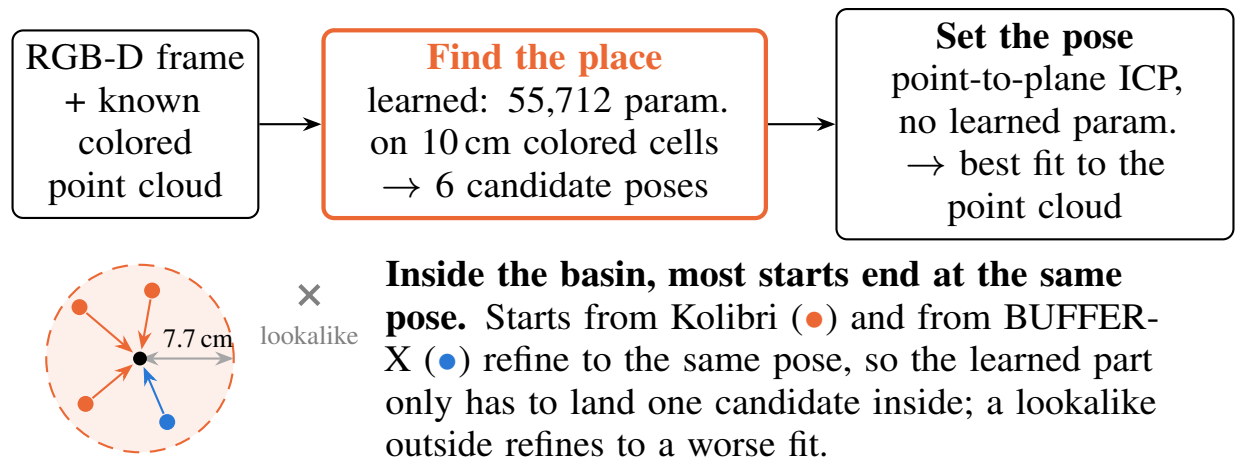
\begin{figure}[t]
\centering
\begin{tikzpicture}[font=\scriptsize, >={Stealth[length=3.5pt]},
  box/.style={draw, rounded corners=2pt, align=center, inner sep=2.5pt, minimum height=1.05cm, anchor=west}]
\node[box, text width=1.45cm] (in) at (0,0) {RGB-D frame\\+ known colored\\point cloud};
\node[box, draw=kolibri, thick, text width=2.75cm] (place) at (2.05,0) {\textbf{\color{kolibri}Find the place}\\learned: \pnum{method.params} param.\\on \pnum{method.cell_cm}\,cm colored cells\\$\rightarrow$ \pnum{method.candidates} candidate poses};
\node[box, text width=2.45cm] (pose) at (5.45,0) {\textbf{Set the pose}\\point-to-plane ICP,\\no learned param.\\$\rightarrow$ best fit to the\\point cloud};
\draw[->] (in) -- (place); \draw[->] (place) -- (pose);
\coordinate (c) at (0.85,-1.55);
\fill[kolibri!10] (c) circle (0.62);
\draw[kolibri, densely dashed] (c) circle (0.62);
\draw[gray!70, <->] (c) -- node[above=-1.5pt, pos=0.62, font=\tiny, text=black] {\pnum{small.basin.e50_cm}\,cm} ($(c)+(0.62,0)$);
\fill (c) circle (1.3pt);
\foreach \px/\py in {-0.40/0.34, 0.08/0.45, -0.36/-0.30} {
  \fill[kolibri] ($(c)+(\px,\py)$) circle (1.5pt);
  \draw[->, kolibri, shorten >=1.7pt, shorten <=1.7pt] ($(c)+(\px,\py)$) -- (c);
}
\fill[bufx] ($(c)+(0.18,-0.42)$) circle (1.5pt);
\draw[->, bufx, shorten >=1.7pt, shorten <=1.7pt] ($(c)+(0.18,-0.42)$) -- (c);
\node[draw=gray, cross out, inner sep=1.4pt, thick] (look) at ($(c)+(1.12,0.42)$) {};
\node[below=0.5pt, font=\tiny, text=gray, align=center] at (look.south) {lookalike};
\node[anchor=west, text width=5.25cm, align=left] at (2.35,-1.55)
 {\textbf{Inside the basin, most starts end at the same pose.} Starts from \ours (\textcolor{kolibri}{$\bullet$}) and from BUFFER-X (\textcolor{bufx}{$\bullet$}) refine to the same pose, so the learned part only has to land one candidate inside; a lookalike outside refines to a worse fit.};
\end{tikzpicture}
\caption[Finding the place is learned, setting the pose is geometry]{\textbf{The network finds candidate places, and geometry refines their poses.}
Point-to-plane iterative closest point (ICP) aligns frame points to nearby surfaces.
The dashed region illustrates the refinement basin.
Its measured boundary is the initial position error at which half the starts succeed within 5\,cm and 5\textdegree{} (\cref{fig:small}a).
A geometric fit score selects the final candidate.}
\label{fig:split}
\end{figure}

\looseness=-1
Our starting point is that finding the place and refining the pose need different levels of accuracy (\cref{fig:split}).
Point-to-plane refinement adjusts a pose by pulling frame points onto nearby surfaces of the point cloud.
It is a variant of iterative closest point (ICP)~\cite{besl1992method,chen1992object}.
Nearby initial poses often converge to the same result, within a region we call the refinement basin.
On real ScanNet frames, \ours and BUFFER-X both start in this region for \pnum{small.basin.both_inside_pct}\% of frames.
Refinement gives the same success or failure outcome on \pnum{small.basin.same_outcome_pct}\% of those frames (\cref{fig:small}a).
The learned component therefore needs to find candidate places close enough for refinement.
Color helps: adding color to Fast Point Feature Histograms (FPFH)~\cite{rusu2009fast}, a geometric descriptor, doubles recall at 5\,cm with our pose estimator (\cref{tab:ablation}).

\looseness=-1
\ours uses a graph network of \pnum{method.params} parameters to describe cells, \pnum{method.cell_cm}\,cm cubes of pooled points, from their color, local shape and neighbors (\cref{fig:method}).
The point cloud is described once and reused.
Groups of consistent cell matches propose candidate poses, refinement adjusts them, and a geometric fit score selects the result.
Color proposes the place and geometry decides, as color aids coarse scene recognition in humans~\cite{oliva2000diagnostic}.

This division of work needs colors that agree between the frame and point cloud up to a change in brightness and contrast.
On gray frames, RGB GeoTransformer, a point cloud registration method with color input, has \pnum{teaser.price.gray_gap} points higher recall than \ours at 5\,cm.
Point cloud registration methods also achieve higher recall in a laser scan colored by a different camera (\cref{sec:price}).

Our contributions are:
\begin{itemize}
\item A division of RGB-D localization into place finding and pose refinement, supported by measurements of when different initial poses give the same refinement outcome.
\item \ours, a small graph network that finds the place from color. With matched refinement, it is competitive with point cloud registration methods on unseen rooms and a held-out test, at a small fraction of their cost.
\item Search without a room label. \ours searches \pnum{building.rooms} rooms within \pnum{building.fast.loss} points of its room-given recall, and a hospital emergency department in \pnum{hosp.fast.ms}\,ms per frame.
\end{itemize}

\section{Related Work}
\label{sec:related}

\paragraph{Point cloud registration.}
\looseness=-1
Point cloud registration methods combine descriptors with an estimator that rejects inconsistent matches~\cite{fischler1981random,yang2020teaser}.
Descriptors progressed from hand-crafted local geometry~\cite{rusu2009fast} to learned representations and matchers~\cite{zeng20173dmatch,huang2021predator,qin2022geometric,yu2023rotation,poiesi2022learning,ao2023buffer}, including methods that transfer across sensors and scales~\cite{seo2025bufferx}.
Colored ICP~\cite{park2017colored} and ColorPCR~\cite{mu2024colorpcr} use color during alignment.
These methods start anew for each pair.
Our reference point cloud is known in advance, so its descriptors can be reused.
The layer exchanging features between neighboring cells, following EdgeConv~\cite{wang2019dgcnn}, only needs to find starting poses for refinement.

\paragraph{Scene-specific relocalization.}
Scene coordinate regression trains a model for each scene to predict the scene position of each image pixel~\cite{shotton2013scene,brachmann2021visual,brachmann2023accelerated}.
It is accurate on 7-Scenes~\cite{shotton2013scene} and 12-Scenes~\cite{valentin2016learning}, where uncertainty in the reference poses limits measurable accuracy~\cite{brachmann2021limits}.

\paragraph{Relocalization in unseen scenes.}
Other methods predict scene coordinates or relative poses without training on the query scene.
They use reference images with known poses or spatial coordinates~\cite{yang2019sanet,tang2021dsm,revaud2023sacreg,dong2025reloc3r}.
MeshLoc~\cite{panek2022meshloc} matches against rendered mesh views, while general-purpose models of scene geometry match image pairs~\cite{wang2024dust3r,leroy2024grounding}.
Our depth input instead supports direct matching to an existing point cloud.
MASt3R and Reloc3r, which relate the query to reference images, have lower recall than \ours when those images are rendered from the point cloud (\cref{sec:main}).

\paragraph{Localization at building scale.}
A common approach retrieves likely places with an image descriptor~\cite{arandjelovic2016netvlad,oquab2024dinov2}, then estimates the pose within them~\cite{sarlin2019coarse}.
Place recognition from laser scans retrieves the matching region of a large point cloud~\cite{kim2018scancontext,uy2018pointnetvlad}.
Image-to-point-cloud methods also use retrieval or restrict the search to regions near known poses~\cite{pham2020lcd,li2021deepi2p,kim2023ep2ploc}.
Systems that estimate scene geometry and camera poses together~\cite{newcombe2011kinectfusion,keetha2025mapanything} solve a different task from localization in a prior scan.
\ours searches all rooms together, without an earlier room decision that can discard the correct place.

\section{Method}
\label{sec:method}

\begin{figure*}[t]
\centering
\includegraphics[width=0.9\linewidth]{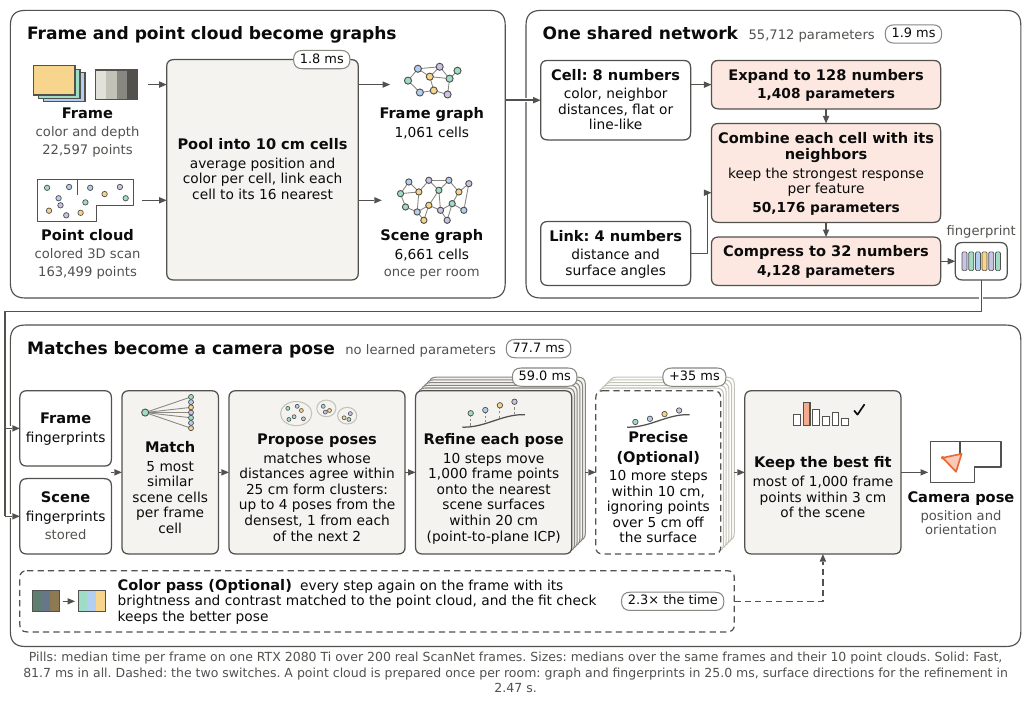}
\caption{\textbf{\ours learns where to start and uses geometry to refine the pose.}
The frame and point cloud become graphs of colored cells.
A shared network assigns each cell a descriptor, or vector for matching.
Consistent matches propose candidate poses, point-to-plane refinement adjusts them, and a geometric fit score selects the result.}
\label{fig:method}
\end{figure*}

\subsection{Task, settings and metric}
\label{sec:task}
The inputs are an RGB-D frame with known camera intrinsics and a colored metric point cloud $\mathcal{S}$ known in advance.
The output is the camera pose $T \in SE(3)$, a rigid rotation and translation that places the frame in $\mathcal{S}$.
With the \emph{room given}, $\mathcal{S}$ contains only the frame's room (\cref{tab:unseen,tab:main}).
With the \emph{room unknown}, it joins rooms or covers a hospital emergency department (\cref{sec:rooms,sec:hospital}).
Recall is the share of frames whose position and rotation errors both fall below a threshold pair: 5\,cm and 5\textdegree, 0.25\,m and 2\textdegree, or 0.5\,m and 5\textdegree.
A frame without a pose counts as a miss.
No evaluation room or scan enters our training.

\ours first describes cells in the frame and point cloud, then groups consistent cell matches into candidate poses, and finally refines and scores those poses (\cref{fig:method}).
Only the cell descriptor is learned.
The point cloud descriptors are computed once and reused.

\subsection{Cell descriptors}
\label{sec:place}
Color distinguishes places even at a coarse scale.
We back-project the frame's depth into space and pool its points into cubic cells of side \pnum{method.cell_cm}\,cm.
Each cell $i$ stores its points' mean position $\mathbf{p}_i$ and color $\mathbf{c}_i$.
Links to its $k=\pnum{method.k}$ nearest cells form the neighborhood $\mathcal{N}(i)$.
The cell normal $\mathbf{n}_i$ points in the direction of least spread among offsets to these neighbors, given by the smallest-eigenvalue eigenvector of their covariance.
The input feature $\mathbf{x}_i$ contains eight values: color centered on mid-gray, the mean and variance of log edge lengths, and linearity, planarity and scattering from the same covariance.
The edge feature $\mathbf{e}_{ij}$ contains its log length and the absolute cosines between the offset and each normal, and between the normals.
Absolute values remove sensitivity to normal orientation.

A message-passing layer~\cite{gilmer2017neural} lets each cell combine its features with those of its neighbors.
It has width \pnum{prose.net.width} and produces the descriptor $\mathbf{f}_i$:
\begin{equation}
\begin{aligned}
\mathbf{h}_i &= \sigma\big(\mathrm{LN}(W_\mathrm{in}\mathbf{x}_i)\big),\\
\mathbf{h}'_i &= \mathrm{LN}\Big(\mathbf{h}_i + \max_{j\in\mathcal{N}(i)} \phi\big([\mathbf{h}_i, \mathbf{h}_j, \mathbf{e}_{ij}]\big)\Big),\\
\mathbf{f}_i &= W_\mathrm{out}\mathbf{h}'_i \,/\, \lVert W_\mathrm{out}\mathbf{h}'_i \rVert .
\end{aligned}
\label{eq:layer}
\end{equation}
Here $\mathbf{h}_i$ and $\mathbf{h}'_i$ are the cell features before and after the update.
$W_\mathrm{in}$ and $W_\mathrm{out}$ are linear layers, $\sigma$ is the Gaussian error linear unit (GELU), $\mathrm{LN}$ is layer normalization, and $\phi$ is a two-layer perceptron.
Brackets concatenate features, and the maximum is taken separately for each feature.
The unit-length descriptor has \pnum{method.desc_dim} values, and the network has \pnum{method.params} parameters.

The same pooling and network describe the reference point cloud.
A graph holds at most \pnum{method.piece_cells} cells.
With the room given, larger rooms use coarser cells until they fit.
With the room unknown, the point cloud is described in pieces at the nominal cell size, each within this limit.
Search then considers the cells of all pieces together.
A frame descriptor $\mathbf{f}_i$ and point cloud descriptor $\mathbf{g}_a$ have similarity $s_{ia}=\mathbf{f}_i^{\top}\mathbf{g}_a/\tau$, where the temperature $\tau=\pnum{prose.net.temperature}$ controls the sharpness of the matching weights.

\subsection{Candidate poses and refinement}
\label{sec:pose}
Each frame cell is matched to its \pnum{prose.est.topk} most similar point cloud cells.
A softmax converts $s_{ia}$ over all point cloud cells into match weights.
Matches $(i,a)$ and $(j,b)$ are consistent if their distances agree:
$\big|\lVert\mathbf{p}_i-\mathbf{p}_j\rVert-\lVert\mathbf{q}_a-\mathbf{q}_b\rVert\big|<\rho$,
where $\mathbf{q}_a$ and $\mathbf{q}_b$ are point cloud cell positions and $\rho=\pnum{prose.est.rho_cm}$\,cm.
The \pnum{prose.est.seeds} matches with the most consistent partners each seed a \emph{consensus set}, an idea from the spatial-consistency estimator SC$^2$-PCR~\cite{chen2022sc2pcr}.
This set contains the starting match and its most consistent partners, one in \pnum{method.consensus_den} of all matches and at least \pnum{method.consensus_min}.
A rigid fit weighted by the match weights gives a candidate pose.

Repeated furniture can produce a large group of consistent matches at the wrong place.
The estimator therefore keeps several distinct candidates.
It retains the \pnum{prose.est.first_round} best first-round candidates separated by at least \pnum{method.distinct_m}\,m or \pnum{method.distinct_deg}\textdegree.
It removes matches explained by these poses, with residuals below \pnum{prose.est.peel_m}\,m, then makes \pnum{prose.est.rounds} further rounds.
Each round adds its best candidate if distinct and supported by at least \pnum{method.min_ratio_pct}\% as many agreeing matches, or inliers, as the best first candidate.
There are at most \pnum{method.candidates} candidates.

Point-to-plane ICP refines each candidate using \pnum{prose.est.icp_points} frame points sampled at even steps through the frame.
It aligns these points to the planes of their nearest point cloud neighbors for \pnum{prose.est.icp_iters} iterations.
The correspondence radius shrinks from \pnum{prose.est.icp_radius_cm}\,cm to \pnum{method.icp_radius_end_cm}\,cm.
The fit score counts how many sampled points lie within \pnum{method.verify_cm}\,cm of the point cloud, and the highest-scoring candidate wins.
Describing the frame takes \pnum{prose.time.describe_ms}\,ms of the total \pnum{time.fast.median_ms}\,ms per frame.
Candidate refinement accounts for most of the remaining time.

\subsection{Operating points}
\label{sec:switches}
The pipeline above is our standard setting, \fast.
Two optional operations trade time for accuracy or tolerance to color changes, without retraining.
\precise adds a tighter second refinement to every candidate.
It uses point-to-plane ICP with Tukey weights, which suppress large residuals, at scale \pnum{prose.second.scale_cm}\,cm and a radius that shrinks from \pnum{prose.second.radius_cm}\,cm.
The \colorpass adds another reading of the frame.
A shared affine transform of its color channels matches its luminance mean and spread to those of the point cloud.
Both the original and adjusted colors go through the full pipeline.
The fit score selects the adjusted reading only if it fits strictly better.
Unless stated otherwise, \ours uses \fast with one color reading.

\subsection{Training}
\label{sec:training}
Training uses \pnum{prose.train.maps} rooms, disjoint from every evaluation set: \pnum{prose.train.scannet_rooms} ScanNet rooms~\cite{dai2017scannet} and \pnum{prose.train.building_maps} spaces in a university building.
Color and depth views rendered along walks provide the training frames.
No real camera frame enters training.
Augmentations include depth noise that grows with squared range, changes in scale, brightness, contrast and white balance, and missing sectors.
For each frame cell, the positive match is the point cloud cell nearest its true position, if within \pnum{method.positive_m}\,m.
The loss is cross-entropy over the softmax matching weights.
Training uses AdamW with learning rate \pnum{prose.train.lr} for \pnum{prose.train.steps} steps of \pnum{prose.train.batch_views} views, taking at most \pnum{prose.train.minutes_max} minutes on an RTX 4090.

\section{Experiments}
\label{sec:exp}

\subsection{Setup}
\label{sec:setup}
\begin{table}[t]
\centering
\scriptsize
\setlength{\tabcolsep}{1.8pt}
\begin{tabular}{@{}lrrll@{}}
\toprule
Set & Frames & Rooms & Point cloud & Role \\
\midrule
\multicolumn{5}{@{}l}{\textit{Room given (\cref{tab:unseen,tab:main}) and rooms joined into one point cloud (\cref{sec:rooms})}} \\
ScanNet-10 & \pnum{prose.frames.scannet} & \pnum{sets.scannet10.rooms} & same recording & dev.: design, ablation rooms \\
ScanNet-10-S2 & \pnum{prose.frames.xsession} & \pnum{sets.xsession.rooms} & first session & development \\
7-Scenes & \pnum{prose.frames.sevenscenes} & \pnum{sets.sevenscenes.rooms} & other recordings & development, ablation \\
12-Scenes & \pnum{prose.frames.twelvescenes} & \pnum{sets.twelvescenes.rooms} & other recordings & development \\
ScanNet-96 & \pnum{campaign_a.frames} & \pnum{campaign_a.rooms} & same recording & test \\
ScanNet-96-S2 & \pnum{campaign_a.rescan.frames} & \pnum{campaign_a.rescan.rooms} & first session & test \\
Held-out test & \pnum{final.real.n} & \pnum{final.rooms} & same recording & test, run once \\
\quad rendered & \pnum{final.rendered.n}$^\dagger$ & \pnum{final.rooms} & same recording & test, run once \\
\midrule
\multicolumn{5}{@{}l}{\textit{Room unknown, whole scans (\cref{sec:hospital,sec:price})}} \\
Emergency dept. & \pnum{hosp.frames}$^\dagger$ & many & handheld laser scan & test \\
Bonn & \pnum{bonn.frames} & one & terrestrial laser scan & scope limit \\
\bottomrule
\end{tabular}
\caption{\textbf{Evaluation sets and their roles.}
A second session (S2) is a later recording localized in the first session's point cloud.
Development sets determine the design and settings.
Test sets play no part in either.
$^\dagger$ marks views rendered from the point cloud with simulated depth noise.
Other frames come from real sensors.}
\label{tab:sets}
\end{table}

\begin{table}[t]
\centering
\scriptsize
\setlength{\tabcolsep}{2.2pt}
\begin{tabular}{@{}l r@{}l r@{}l r@{}l r@{}l@{}}
\toprule
 & \multicolumn{4}{c}{ScanNet-96 (7,986 frames)} & \multicolumn{4}{c@{}}{ScanNet-96-S2 (5,133 frames)} \\
\cmidrule(lr){2-5}\cmidrule(l){6-9}
 & \multicolumn{2}{c}{\fast} & \multicolumn{2}{c}{\precise} & \multicolumn{2}{c}{\fast} & \multicolumn{2}{c@{}}{\precise} \\
\midrule
\oursrow \ours & 81.4\,{\tiny$\pm$1.8} &  & 85.3\,{\tiny$\pm$1.6} &  & 65.1\,{\tiny$\pm$3.4} &  & 66.5\,{\tiny$\pm$3.4} &\oursend  \\
\oursrow \quad + \colorpass & \best{81.6}\,{\tiny$\pm$1.8} &  & \best{85.5}\,{\tiny$\pm$1.6} &  & \best{65.5}\,{\tiny$\pm$3.4} &  & 67.0\,{\tiny$\pm$3.3} &\oursend  \\
\midrule
BUFFER-X & 69.4\,{\tiny$\pm$2.8} & $^{\blacktriangle}$ & 72.4\,{\tiny$\pm$2.8} & $^{\blacktriangle}$ & 61.5\,{\tiny$\pm$3.7} & $^{\blacktriangle}$ & 62.7\,{\tiny$\pm$3.6} & $^{\blacktriangle}$ \\
BUFFER-X-Lite & 70.4\,{\tiny$\pm$2.7} & $^{\blacktriangle}$ & 73.3\,{\tiny$\pm$2.7} & $^{\blacktriangle}$ & 61.9\,{\tiny$\pm$3.7} & $^{\blacktriangle}$ & 63.0\,{\tiny$\pm$3.6} & $^{\blacktriangle}$ \\
GeDi & 80.3\,{\tiny$\pm$1.9} & $^{\approx\blacktriangle}$ & 84.2\,{\tiny$\pm$1.6} & $^{\approx\blacktriangle}$ & \best{65.5}\,{\tiny$\pm$3.3} & $^{\approx}$ & \best{67.1}\,{\tiny$\pm$3.3} & $^{\approx}$ \\
RGB GeoTransformer & 77.8\,{\tiny$\pm$1.9} & $^{\blacktriangle}$ & 81.4\,{\tiny$\pm$1.8} & $^{\blacktriangle}$ & 64.0\,{\tiny$\pm$3.3} & $^{\approx\blacktriangle}$ & 65.7\,{\tiny$\pm$3.4} & $^{\approx}$ \\
ColorPCR & 80.3\,{\tiny$\pm$1.9} & $^{\approx\blacktriangle}$ & 84.1\,{\tiny$\pm$1.7} & $^{\approx\blacktriangle}$ & 63.2\,{\tiny$\pm$3.4} & $^{\approx\blacktriangle}$ & 64.4\,{\tiny$\pm$3.5} & $^{\approx\blacktriangle}$ \\
RoITr & 75.2\,{\tiny$\pm$2.2} & $^{\blacktriangle}$ & 78.4\,{\tiny$\pm$2.1} & $^{\blacktriangle}$ & 64.1\,{\tiny$\pm$3.4} & $^{\approx}$ & 65.7\,{\tiny$\pm$3.3} & $^{\approx}$ \\
PREDATOR & 49.6\,{\tiny$\pm$2.5} & $^{\blacktriangle}$ & 51.8\,{\tiny$\pm$2.4} & $^{\blacktriangle}$ & 45.6\,{\tiny$\pm$3.4} & $^{\blacktriangle}$ & 46.7\,{\tiny$\pm$3.3} & $^{\blacktriangle}$ \\
FPFH + TEASER++ & 33.2\,{\tiny$\pm$2.4} & $^{\blacktriangle}$ & 34.7\,{\tiny$\pm$2.4} & $^{\blacktriangle}$ & 29.0\,{\tiny$\pm$3.2} & $^{\blacktriangle}$ & 29.8\,{\tiny$\pm$3.2} & $^{\blacktriangle}$ \\
\bottomrule
\end{tabular}

\caption{\textbf{Room-given test on ScanNet-96 and its later recording sessions.}
Recall (\%) uses 5\,cm and 5\textdegree.
Here and in later tables, gray rows are \ours and bold marks the best value per column.
\ours averages three training seeds.
Baselines use one run and refinement matched to \fast or \precise.
The $\pm$ entries give interval half-widths over walks.
Marks compare \ours with one color reading against each cell, using the criteria in \cref{sec:setup}: $\approx$ equivalent, $\blacktriangle$ \ours leads, $\approx\blacktriangle$ both.}
\label{tab:unseen}
\end{table}

\paragraph{Evaluation sets.}
\Cref{tab:sets} names the evaluation sets and their roles.
Reference camera poses come from RGB-D reconstruction~\cite{dai2017scannet,brachmann2021limits}.
For ScanNet~\cite{dai2017scannet}, the point cloud and frames come from the same recording.
A \emph{second session} (S2) is a later recording localized in the first session's point cloud.
The 7-Scenes~\cite{shotton2013scene} and 12-Scenes~\cite{valentin2016learning} point clouds come from their training recordings, so their queries also cross sessions.
Development uses ScanNet-10, ScanNet-10-S2, every tenth 7-Scenes test frame, and 12-Scenes.
Five ScanNet-10 \emph{design rooms} determine the network and settings.
The other five \emph{ablation rooms}, together with 7-Scenes, support the component comparisons in \cref{sec:small}.
ScanNet-96 contains the \pnum{campaign_a.rooms} of the next \pnum{campaign_a.rooms_checked} ScanNet rooms whose reference poses fit their point clouds.
ScanNet-96-S2 contains \pnum{campaign_a.rescan.scans} later scans of \pnum{campaign_a.rescan.rooms} of these rooms.
Neither set shares a room with training.
A further \pnum{final.rooms} rooms form the held-out test, evaluated once after the method was frozen.

\paragraph{Baselines.}
The point cloud registration methods are BUFFER-X~\cite{seo2025bufferx} and its released light setting BUFFER-X-Lite, GeDi~\cite{poiesi2022learning}, ColorPCR~\cite{mu2024colorpcr}, RoITr~\cite{yu2023rotation}, PREDATOR~\cite{huang2021predator}, FPFH with the robust estimator TEASER++~\cite{rusu2009fast,yang2020teaser}, and RGB GeoTransformer.
The latter adds color input to GeoTransformer~\cite{qin2022geometric} and is fine-tuned on our training rooms.
The other learned methods use released weights trained on 3DMatch~\cite{zeng20173dmatch} or its colored version~\cite{mu2024colorpcr}.
3DMatch contains 7-Scenes scenes and \pnum{tab_main.threedmatch.twelvescenes_rooms} of the \pnum{tab_main.twelvescenes.rooms} 12-Scenes rooms, which the colored version omits.
We report each baseline as released (\emph{native}) and with our refinement (\emph{+ refinement}).
BUFFER-X retains its own refinement in both cases.
Comparisons with \precise give baselines the same second refinement stage.
BUFFER-X chooses its patch radius from the larger input point cloud and samples a fixed number of keypoints, which suits similarly sized fragments.
It therefore receives single rooms or \pnum{hosp.tile_m}\,m cubes, never a point cloud spanning several rooms.

\paragraph{Statistics and timing.}
A \emph{walk} is a run of consecutive frames.
Recall differences carry 95\% intervals from a bootstrap that resamples whole walks, pairs methods on the same frames, and averages over three training seeds.
The $\pm$ entries in \cref{tab:unseen,tab:main,tab:points} give interval half-widths.
Methods are \emph{equivalent} when the 90\% interval lies within $\pm$\pnum{tab_main.margin} recall points.
A method \emph{leads} when the 95\% interval excludes zero in its favor.
Equivalence and a lead can both hold.
A comparison meeting neither criterion is \emph{inconclusive}.
Resampling rooms or seeds gives the same verdicts, and the largest standard deviation of \ours recall across seeds is \pnum{tables.seed_sd_max} points.
Unless stated otherwise, time and memory are medians over \pnum{tab_main.timing_frames} fixed ScanNet-10 frames, processed individually on four CPU threads and an RTX 2080 Ti.
\ours and GeDi reuse point cloud descriptors.
Other baselines run as released.

\subsection{Room given: accuracy and cost}
\label{sec:main}
\begin{table*}[t]
\centering
\scriptsize
\setlength{\tabcolsep}{1.25pt}
\begin{tabular}{@{}l r c r r r r@{}l r@{}l r@{}l r@{}l r@{}l r@{}l r@{}l r@{}l@{}}
\toprule
 & & Test scenes & \multicolumn{3}{c}{Per frame} & \multicolumn{4}{c}{\begin{tabular}{@{}c@{}}ScanNet-10\\871 frames\end{tabular}} & \multicolumn{4}{c}{\begin{tabular}{@{}c@{}}7-Scenes\\1,700 frames\end{tabular}} & \multicolumn{4}{c}{\begin{tabular}{@{}c@{}}12-Scenes\\575 frames\end{tabular}} & \multicolumn{4}{c}{\begin{tabular}{@{}c@{}}ScanNet-10-S2\\683 frames\end{tabular}} \\
\cmidrule(lr){4-6}\cmidrule(lr){7-10}\cmidrule(lr){11-14}\cmidrule(lr){15-18}\cmidrule(l){19-22}
Method & Parameters & in training & median ms & p95 ms & GB & \multicolumn{2}{c}{5\,cm} & \multicolumn{2}{c}{0.5\,m} & \multicolumn{2}{c}{5\,cm} & \multicolumn{2}{c}{0.5\,m} & \multicolumn{2}{c}{5\,cm} & \multicolumn{2}{c}{0.5\,m} & \multicolumn{2}{c}{5\,cm} & \multicolumn{2}{c}{0.5\,m} \\
\midrule
\oursrow \ours \fast & 55.7k & none & \best{81.7} & \best{251.1} & \best{1.44} & 75.7{\tiny$\pm$6.5} &  & 94.9{\tiny$\pm$2.5} &  & 93.0{\tiny$\pm$2.5} &  & 98.5{\tiny$\pm$0.9} &  & 97.9{\tiny$\pm$3.1} &  & 99.7{\tiny$\pm$0.6} &  & 73.3{\tiny$\pm$7.4} &  & 95.3{\tiny$\pm$3.3} &\oursend  \\
\oursrow \ours \precise & 55.7k & none & 116.4 & 417.4 & \best{1.44} & \best{79.2}{\tiny$\pm$5.9} &  & \best{95.1}{\tiny$\pm$2.5} &  & \best{93.1}{\tiny$\pm$2.7} &  & \best{98.6}{\tiny$\pm$0.9} &  & 98.6{\tiny$\pm$2.0} &  & 99.7{\tiny$\pm$0.6} &  & \best{75.9}{\tiny$\pm$7.3} &  & 95.4{\tiny$\pm$3.4} &\oursend  \\
\midrule
BUFFER-X~\cite{seo2025bufferx} & 0.91M & 7S, 12S & 1,416 & 1,922 & 6.21 & 66.4{\tiny$\pm$6.7} & $^{\blacktriangle}$ & 91.3{\tiny$\pm$3.8} & $^{\blacktriangle}$ & 72.7{\tiny$\pm$7.8} & $^{\blacktriangle}$ & 96.2{\tiny$\pm$2.5} & $^{\blacktriangle}$ & 82.1{\tiny$\pm$7.8} & $^{\blacktriangle}$ & 91.8{\tiny$\pm$5.0} & $^{\blacktriangle}$ & 66.9{\tiny$\pm$8.6} & $^{\blacktriangle}$ & 95.2{\tiny$\pm$2.9} & $^{\circ}$ \\
\quad + refinement & &  & 1,444 & 1,948 & 6.21 & 74.7{\tiny$\pm$6.9} & $^{\approx}$ & 92.9{\tiny$\pm$3.6} & $^{\circ}$ & 92.2{\tiny$\pm$3.6} & $^{\approx}$ & 97.5{\tiny$\pm$2.3} & $^{\circ}$ & 91.8{\tiny$\pm$4.7} & $^{\blacktriangle}$ & 93.4{\tiny$\pm$4.4} & $^{\blacktriangle}$ & 73.8{\tiny$\pm$7.3} & $^{\approx}$ & 97.1{\tiny$\pm$2.0} & $^{\circ}$ \\
\midrule
BUFFER-X-Lite~\cite{seo2025bufferx} & 0.91M & 7S, 12S & 1,280 & 1,896 & 6.21 & 63.9{\tiny$\pm$6.7} & $^{\blacktriangle}$ & 90.7{\tiny$\pm$4.2} & $^{\blacktriangle}$ & 66.6{\tiny$\pm$7.3} & $^{\blacktriangle}$ & 96.4{\tiny$\pm$2.5} & $^{\blacktriangle}$ & 81.6{\tiny$\pm$7.9} & $^{\blacktriangle}$ & 91.7{\tiny$\pm$5.6} & $^{\blacktriangle}$ & 62.1{\tiny$\pm$9.6} & $^{\blacktriangle}$ & 95.0{\tiny$\pm$2.9} & $^{\circ}$ \\
\quad + refinement & &  & 1,299 & 1,925 & 6.21 & 74.4{\tiny$\pm$6.8} & $^{\approx}$ & 92.9{\tiny$\pm$3.5} & $^{\circ}$ & 92.3{\tiny$\pm$3.5} & $^{\approx}$ & 97.5{\tiny$\pm$2.3} & $^{\circ}$ & 91.7{\tiny$\pm$4.7} & $^{\blacktriangle}$ & 93.4{\tiny$\pm$4.5} & $^{\blacktriangle}$ & 73.5{\tiny$\pm$7.0} & $^{\approx}$ & 96.8{\tiny$\pm$2.2} & $^{\circ}$ \\
\midrule
GeDi~\cite{poiesi2022learning} & 3.11M & 7S, 12S & 24,149 & 43,869 & 5.27 & 78.8{\tiny$\pm$5.8} & $^{\circ}$ & 92.8{\tiny$\pm$3.5} & $^{\circ}$ & 70.8{\tiny$\pm$9.2} & $^{\blacktriangle}$ & 90.8{\tiny$\pm$3.9} & $^{\blacktriangle}$ & 98.8{\tiny$\pm$0.9} & $^{\circ}$ & 99.7{\tiny$\pm$0.4} & $^{\approx}$ & 74.4{\tiny$\pm$6.8} & $^{\circ}$ & 95.9{\tiny$\pm$2.5} & $^{\circ}$ \\
\quad + refinement & &  & 24,179 & 43,902 & 5.27 & 76.2{\tiny$\pm$7.1} & $^{\approx}$ & \best{95.1}{\tiny$\pm$2.6} & $^{\approx}$ & 90.9{\tiny$\pm$3.8} & $^{\circ}$ & 95.8{\tiny$\pm$2.4} & $^{\blacktriangle}$ & 98.1{\tiny$\pm$2.9} & $^{\approx}$ & 99.7{\tiny$\pm$0.6} & $^{\approx}$ & 73.6{\tiny$\pm$7.1} & $^{\approx}$ & \best{97.5}{\tiny$\pm$1.7} & $^{\circ}$ \\
\midrule
RGB GeoTransformer~\cite{qin2022geometric} & 9.83M & 7S, 12S & 2,501 & 6,338 & 2.70 & 76.3{\tiny$\pm$5.5} & $^{\circ}$ & 88.2{\tiny$\pm$4.4} & $^{\blacktriangle}$ & 66.1{\tiny$\pm$9.8} & $^{\blacktriangle}$ & 92.4{\tiny$\pm$3.2} & $^{\blacktriangle}$ & \best{99.1}{\tiny$\pm$0.8} & $^{\circ}$ & \best{99.8}{\tiny$\pm$0.2} & $^{\approx}$ & 72.6{\tiny$\pm$7.2} & $^{\circ}$ & 93.0{\tiny$\pm$3.4} & $^{\circ}$ \\
\quad + refinement & &  & 2,530 & 6,372 & 2.70 & 73.5{\tiny$\pm$7.4} & $^{\blacktriangle}$ & 89.4{\tiny$\pm$4.0} & $^{\blacktriangle}$ & 89.8{\tiny$\pm$4.3} & $^{\blacktriangle}$ & 94.9{\tiny$\pm$2.9} & $^{\blacktriangle}$ & 98.1{\tiny$\pm$2.9} & $^{\approx}$ & 99.7{\tiny$\pm$0.4} & $^{\approx}$ & 72.3{\tiny$\pm$7.1} & $^{\approx}$ & 94.3{\tiny$\pm$2.9} & $^{\circ}$ \\
\midrule
ColorPCR~\cite{mu2024colorpcr} & 10.03M & 7S & 4,077 & 10,878 & 3.90 & 66.4{\tiny$\pm$6.8} & $^{\blacktriangle}$ & 89.1{\tiny$\pm$4.5} & $^{\blacktriangle}$ & 63.5{\tiny$\pm$8.8} & $^{\blacktriangle}$ & 89.1{\tiny$\pm$4.8} & $^{\blacktriangle}$ & 95.5{\tiny$\pm$2.3} & $^{\circ}$ & 98.4{\tiny$\pm$1.0} & $^{\approx\blacktriangle}$ & 58.6{\tiny$\pm$9.8} & $^{\blacktriangle}$ & 81.7{\tiny$\pm$7.6} & $^{\blacktriangle}$ \\
\quad + refinement & &  & 4,107 & 10,913 & 3.90 & 72.6{\tiny$\pm$7.0} & $^{\blacktriangle}$ & 91.4{\tiny$\pm$3.6} & $^{\blacktriangle}$ & 86.3{\tiny$\pm$5.0} & $^{\blacktriangle}$ & 92.4{\tiny$\pm$4.0} & $^{\blacktriangle}$ & 97.2{\tiny$\pm$2.9} & $^{\approx}$ & 98.8{\tiny$\pm$0.9} & $^{\approx}$ & 67.2{\tiny$\pm$8.7} & $^{\blacktriangle}$ & 86.1{\tiny$\pm$6.7} & $^{\blacktriangle}$ \\
\midrule
RoITr~\cite{yu2023rotation} & 10.10M & 7S, 12S & 6,402 & 22,324 & 2.07 & 60.2{\tiny$\pm$6.9} & $^{\blacktriangle}$ & 86.3{\tiny$\pm$4.5} & $^{\blacktriangle}$ & 55.8{\tiny$\pm$13.3} & $^{\blacktriangle}$ & 81.2{\tiny$\pm$10.6} & $^{\blacktriangle}$ & 77.2{\tiny$\pm$8.9} & $^{\blacktriangle}$ & 92.5{\tiny$\pm$5.8} & $^{\blacktriangle}$ & 61.5{\tiny$\pm$6.9} & $^{\blacktriangle}$ & 90.6{\tiny$\pm$4.0} & $^{\circ}$ \\
\quad + refinement & &  & 6,429 & 22,353 & 2.07 & 72.1{\tiny$\pm$7.3} & $^{\blacktriangle}$ & 89.2{\tiny$\pm$4.0} & $^{\blacktriangle}$ & 81.4{\tiny$\pm$9.1} & $^{\blacktriangle}$ & 86.1{\tiny$\pm$9.0} & $^{\blacktriangle}$ & 92.9{\tiny$\pm$6.8} & $^{\blacktriangle}$ & 94.4{\tiny$\pm$4.9} & $^{\blacktriangle}$ & 71.4{\tiny$\pm$7.3} & $^{\circ}$ & 93.6{\tiny$\pm$3.5} & $^{\circ}$ \\
\midrule
PREDATOR~\cite{huang2021predator} & 7.43M & 7S, 12S & 728.1 & 1,287 & 3.88 & 13.4{\tiny$\pm$4.3} & $^{\blacktriangle}$ & 38.2{\tiny$\pm$6.9} & $^{\blacktriangle}$ & 14.2{\tiny$\pm$8.1} & $^{\blacktriangle}$ & 33.7{\tiny$\pm$12.2} & $^{\blacktriangle}$ & 30.4{\tiny$\pm$8.8} & $^{\blacktriangle}$ & 52.9{\tiny$\pm$10.6} & $^{\blacktriangle}$ & 20.6{\tiny$\pm$6.6} & $^{\blacktriangle}$ & 46.9{\tiny$\pm$7.7} & $^{\blacktriangle}$ \\
\quad + refinement & &  & 754.0 & 1,316 & 3.88 & 47.4{\tiny$\pm$7.6} & $^{\blacktriangle}$ & 53.4{\tiny$\pm$7.0} & $^{\blacktriangle}$ & 47.0{\tiny$\pm$13.3} & $^{\blacktriangle}$ & 51.4{\tiny$\pm$12.8} & $^{\blacktriangle}$ & 72.5{\tiny$\pm$8.5} & $^{\blacktriangle}$ & 74.4{\tiny$\pm$9.0} & $^{\blacktriangle}$ & 53.1{\tiny$\pm$7.9} & $^{\blacktriangle}$ & 67.6{\tiny$\pm$6.6} & $^{\blacktriangle}$ \\
\midrule
FPFH + TEASER++~\cite{rusu2009fast,yang2020teaser} & none & none & 215.0 & 648.5 & -- & 11.0{\tiny$\pm$3.9} & $^{\blacktriangle}$ & 24.9{\tiny$\pm$6.7} & $^{\blacktriangle}$ & 2.6{\tiny$\pm$1.5} & $^{\blacktriangle}$ & 10.8{\tiny$\pm$4.5} & $^{\blacktriangle}$ & 8.0{\tiny$\pm$2.9} & $^{\blacktriangle}$ & 16.3{\tiny$\pm$5.0} & $^{\blacktriangle}$ & 9.5{\tiny$\pm$3.4} & $^{\blacktriangle}$ & 29.6{\tiny$\pm$8.0} & $^{\blacktriangle}$ \\
\quad + refinement & &  & 236.9 & 701.0 & -- & 30.7{\tiny$\pm$7.4} & $^{\blacktriangle}$ & 35.2{\tiny$\pm$7.2} & $^{\blacktriangle}$ & 19.5{\tiny$\pm$6.3} & $^{\blacktriangle}$ & 20.6{\tiny$\pm$6.5} & $^{\blacktriangle}$ & 31.3{\tiny$\pm$7.3} & $^{\blacktriangle}$ & 32.3{\tiny$\pm$6.9} & $^{\blacktriangle}$ & 33.7{\tiny$\pm$8.0} & $^{\blacktriangle}$ & 45.7{\tiny$\pm$9.1} & $^{\blacktriangle}$ \\
\midrule
\multicolumn{3}{@{}l}{\textit{Trained on each test scene}} & \multicolumn{3}{c}{\textit{training, per frame}} & \multicolumn{16}{l}{} \\
ACE~\cite{brachmann2023accelerated} (RGB) & 2.1M + 5.6M & our run & \multicolumn{3}{c}{258\,s (291\,s$^\dagger$), 29\,ms} & -- &  & -- &  & 76.8{\tiny$\pm$8.1} &  & 99.4{\tiny$\pm$0.6} &  & 99.5{\tiny$\pm$0.6} &  & 100.0{\tiny$\pm$0.0} &  & -- &  & -- &  \\
DSAC*~\cite{brachmann2021visual} (RGB) & 28\,MB$^\dagger$ & from~\cite{brachmann2023accelerated,brachmann2021visual} & \multicolumn{3}{c}{15\,h$^\dagger$, 75\,ms$^\dagger$} & -- &  & -- &  & 84.0 & $^\dagger$ & -- &  & 99.2 & $^\dagger$ & -- &  & -- &  & -- &  \\
DSAC*~\cite{brachmann2021visual} (RGB-D) & 28\,MB$^\dagger$ & from~\cite{brachmann2021limits,brachmann2021visual} & \multicolumn{3}{c}{2.5\,d$^\dagger$} & -- &  & -- &  & 94.7 & $^\dagger$ & -- &  & -- &  & -- &  & -- &  & -- &  \\
\bottomrule
\end{tabular}

\caption{\textbf{Room-given development sets: recall and cost.}
Recall (\%) uses 5\,cm and 5\textdegree{} or 0.5\,m and 5\textdegree.
\ours averages three training seeds, and each baseline uses one run.
The $\pm$ entries give interval half-widths over walks.
Marks compare \fast with each row under the criteria in \cref{sec:setup}: $\approx$ equivalent, $\blacktriangle$ \ours leads, $\circ$ inconclusive, $\approx\blacktriangle$ both.
Results with the second refinement stage are in the supplement.
Time and memory are medians over \pnum{tab_main.timing_frames} ScanNet-10 frames on an RTX 2080 Ti (for ACE, over its scenes).
FPFH + TEASER++ uses the CPU.
The third column identifies test scenes present in training data, or the source of a row in the last block.
Methods in the last block train on each test scene and are never bold, and $^\dagger$ marks values from the cited papers, on their full test splits and the GPUs they name.}
\label{tab:main}
\end{table*}

\begin{figure}[t]
\centering
\includegraphics[width=0.92\linewidth]{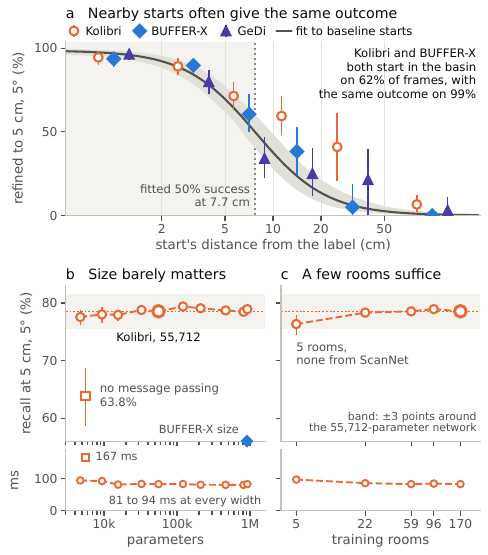}
\caption{\textbf{Refinement makes a small descriptor sufficient.}
(a)~Success within 5\,cm and 5\textdegree{} on ScanNet-10 as a function of initial position error.
The curve fits BUFFER-X and GeDi starts, and the dashed line marks the distance where half succeed.
(b)~Recall at the same threshold on the ablation rooms and median time as network size changes.
Bars show paired intervals, and the band shows the equivalence margin.
(c)~Recall as the number of training rooms changes.}
\label{fig:small}
\end{figure}

\begin{table}[t]
\centering
\scriptsize
\setlength{\tabcolsep}{1.6pt}
\begin{tabular}{@{}l r rr rr@{}}
\toprule
 & & \multicolumn{2}{c}{\fast} & \multicolumn{2}{c@{}}{\precise} \\
\cmidrule(lr){3-4}\cmidrule(l){5-6}
& Params & ScanNet & 7-Scenes & ScanNet & 7-Scenes \\
\midrule
\oursrow Full network & 55.7k & 78.5 & \best{93.0} & 80.1 &\oursend \best{93.1} \\
\midrule
\multicolumn{6}{@{}l}{\textit{One change to the network, retrained}} \\
No message passing & 5.5k & \textminus{}14.6 & \textminus{}24.4 & \textminus{}13.4 & \textminus{}22.7 \\[-1pt]
 & & {\tiny[\textminus{}20.9, \textminus{}8.5]} & {\tiny[\textminus{}32.5, \textminus{}16.7]} & {\tiny[\textminus{}18.9, \textminus{}8.4]} & {\tiny[\textminus{}31.1, \textminus{}15.2]} \\
Geometry only & 55.3k & \textminus{}26.2 & \textminus{}72.7 & \textminus{}26.2 & \textminus{}72.0 \\[-1pt]
 & & {\tiny[\textminus{}33.5, \textminus{}19.3]} & {\tiny[\textminus{}79.4, \textminus{}65.9]} & {\tiny[\textminus{}33.3, \textminus{}19.1]} & {\tiny[\textminus{}78.5, \textminus{}65.2]} \\
Color only & 54.7k & \textminus{}3.3 & \textminus{}0.8 & \textminus{}3.4 & \textminus{}0.9 \\[-1pt]
 & & {\tiny[\textminus{}5.5, \textminus{}1.3]} & {\tiny[\textminus{}1.6, \textminus{}0.1]} & {\tiny[\textminus{}5.8, \textminus{}1.4]} & {\tiny[\textminus{}1.6, \textminus{}0.2]} \\
Simple node features & 54.9k & \textminus{}2.5 & \textminus{}0.5 & \textminus{}2.0 & \textminus{}0.3 \\[-1pt]
 & & {\tiny[\textminus{}4.3, \textminus{}1.1]} & {\tiny[\textminus{}1.1, +0.1]} & {\tiny[\textminus{}3.7, \textminus{}0.5]} & {\tiny[\textminus{}1.1, +0.4]} \\
Two compact layers & 28.4k & +0.2 & \textminus{}0.9 & +0.8 & \textminus{}0.6 \\[-1pt]
 & & {\tiny[\textminus{}0.9, +1.2]} & {\tiny[\textminus{}2.4, +0.2]} & {\tiny[\textminus{}0.4, +2.3]} & {\tiny[\textminus{}1.9, +0.3]} \\
22 training rooms & 55.7k & \textminus{}0.2 & \textminus{}0.3 & +0.2 & \textminus{}0.1 \\[-1pt]
 & & {\tiny[\textminus{}1.3, +0.7]} & {\tiny[\textminus{}0.9, +0.2]} & {\tiny[\textminus{}1.0, +1.1]} & {\tiny[\textminus{}0.5, +0.3]} \\
\midrule
\multicolumn{6}{@{}l}{\textit{Another descriptor, same estimator and refinement}} \\
GeDi & 3.11M & \best{+0.9} & \textminus{}0.2 & \best{+0.9} & \textminus{}0.3 \\[-1pt]
 & & {\tiny[\textminus{}1.5, +3.3]} & {\tiny[\textminus{}1.8, +1.3]} & {\tiny[\textminus{}1.1, +3.4]} & {\tiny[\textminus{}1.6, +1.1]} \\
RGB GeoTransformer & 9.83M & +0.2 & \textminus{}2.5 & \best{+0.9} & \textminus{}2.8 \\[-1pt]
 & & {\tiny[\textminus{}1.8, +2.6]} & {\tiny[\textminus{}5.1, \textminus{}0.1]} & {\tiny[\textminus{}0.9, +2.9]} & {\tiny[\textminus{}5.4, \textminus{}0.5]} \\
FPFH & 0 & \textminus{}48.3 & \textminus{}71.6 & \textminus{}48.3 & \textminus{}70.4 \\[-1pt]
 & & {\tiny[\textminus{}58.5, \textminus{}38.4]} & {\tiny[\textminus{}78.8, \textminus{}64.1]} & {\tiny[\textminus{}57.9, \textminus{}38.4]} & {\tiny[\textminus{}77.9, \textminus{}62.7]} \\
FPFH + color & 0 & \textminus{}16.7 & \textminus{}19.6 & \textminus{}14.9 & \textminus{}18.7 \\[-1pt]
 & & {\tiny[\textminus{}24.0, \textminus{}10.3]} & {\tiny[\textminus{}26.4, \textminus{}13.6]} & {\tiny[\textminus{}22.2, \textminus{}8.3]} & {\tiny[\textminus{}25.0, \textminus{}12.8]} \\
Ours, untrained & 55.7k & \textminus{}15.9 & \textminus{}21.9 & \textminus{}15.2 & \textminus{}20.2 \\[-1pt]
 & & {\tiny[\textminus{}22.8, \textminus{}9.0]} & {\tiny[\textminus{}27.6, \textminus{}16.2]} & {\tiny[\textminus{}22.2, \textminus{}8.4]} & {\tiny[\textminus{}25.7, \textminus{}14.8]} \\
\bottomrule
\end{tabular}

\caption{\textbf{Color and information from neighboring cells carry the result.}
The full network row gives recall (\%) at 5\,cm and 5\textdegree{} on the ScanNet-10 ablation rooms and 7-Scenes.
Other rows give changes in recall points, with paired intervals over walks.
For bold, the full network counts as no change.
Each network change uses three training seeds.}
\label{tab:ablation}
\end{table}

\begin{figure}[t]
\centering
\includegraphics[width=0.88\linewidth]{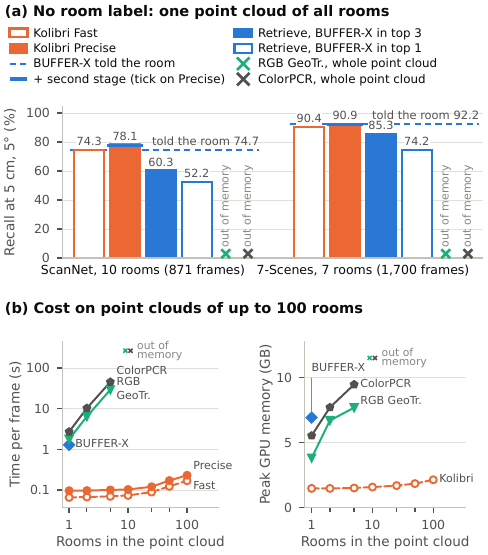}
\caption{\textbf{Searching all rooms avoids losses from choosing a room first.}
(a)~Recall at 5\,cm and 5\textdegree{} in the ten-room and seven-room point clouds, without a room label.
BUFFER-X receives only the true room or retrieved rooms, because its patch radius depends on point cloud size.
(b)~Median time per frame and peak GPU memory as the point cloud grows, on \pnum{building.timing_frames} ScanNet-96 frames and an RTX 2080 Ti.}
\label{fig:rooms}
\end{figure}

With the room given and refinement matched, \fast localizes \pnum{campaign_a.fast.r5}\% of ScanNet-96 frames within 5\,cm and 5\textdegree, against GeDi's \pnum{campaign_a.gedi.r5}\% (\cref{tab:unseen}).
The methods are equivalent at this threshold, and \fast leads BUFFER-X.
Across both ScanNet-96 sets, all four operating points and all three thresholds, none of the \pnum{campaign_a.matched.cells} matched-refinement comparisons shows a significant deficit for \ours.
Development-set comparisons are also equivalent, in favor of \ours, or inconclusive (\cref{tab:main}).
\fast takes \pnum{time.fast.median_ms}\,ms per frame, while BUFFER-X takes \pnum{speedup.bufferx_over_fast} times as long.
It also uses less GPU memory than BUFFER-X and GeDi.
Even after varying their settings, no baseline reaches \fast's ScanNet-10 recall in less than \pnum{teaser.reach.ratio} times its time (\cref{fig:teaser}b).

The held-out test supports the same result on rooms unused in any design choice.
On its sensor frames, \precise reaches \pnum{final.real.precise.r5}\% within 5\,cm and 5\textdegree, against \pnum{final.real.precise_vs_gedi2.base_r5}\% for GeDi with the same refinement stages.
They are equivalent, as are \fast and GeDi with the shared refinement at this threshold.
On rendered views, \precise is equivalent to ColorPCR within 0.5\,m and 5\textdegree{} and leads GeDi and BUFFER-X.

For a different tradeoff, the scene-specific method ACE~\cite{brachmann2023accelerated} first trains for minutes on each test scene.
It reaches \pnum{ref.ace.run.sevenscenes_r5}\% on our 7-Scenes frames within 5\,cm, against \pnum{prose.t1.fast.sevenscenes_r5}\% for \fast (\cref{tab:main}).
The image-based methods MASt3R~\cite{leroy2024grounding} and Reloc3r~\cite{dong2025reloc3r} instead use \pnum{prose.rooms.retrieval_views} rendered views per room.
With the same refinement, they trail \fast on every evaluated set and threshold.

\subsection{Why a small network is enough}
\label{sec:small}
Refinement reduces the accuracy needed from the learned descriptor.
The success rate falls to half at an initial position error of \pnum{small.basin.e50_cm}\,cm (\cref{fig:small}a), which defines the measured basin boundary.
Even when \fast's best initial candidate lies outside this basin, another candidate rescues \pnum{prose.basin.outside_ours}\% of frames, against \pnum{prose.basin.outside_bufferx}\% for BUFFER-X.
Increasing capacity has little effect over the tested range.
From \pnum{small.size.params_min} to \pnum{small.size.params_max} parameters, recall stays within \pnum{small.size.max_abs_diff} point of our network on the ablation rooms (\cref{fig:small}b).
Training on \pnum{small.maps.second} rooms also gives recall equivalent to training on all \pnum{small.maps.max} (\cref{fig:small}c).

What matters is the information exchanged between cells (\cref{tab:ablation}).
With \precise on the ablation rooms, removing message passing costs \pnum{tab_ablation.precise.no_message_passing_loss} points, and removing color costs \pnum{tab_ablation.precise.no_color_loss}.
Keeping only color on the same graph loses \pnum{tab_ablation.precise.color_only_loss} points, so local geometry still matters.
GeDi's descriptor uses \pnum{tab_ablation.gedi_param_ratio} times as many parameters and localizes more frames before refinement.
That advantage disappears after refinement.

\subsection{Search without a room label}
\label{sec:rooms}
\begin{table}[t]
\centering
\scriptsize
\setlength{\tabcolsep}{4pt}
\def\tabpointsdash{}
\begin{tabular}{@{}l r@{}l r@{}l r@{}l r@{}l@{}}
\toprule
 & \multicolumn{4}{c}{One color reading} & \multicolumn{4}{c@{}}{Color pass} \\
\cmidrule(lr){2-5}\cmidrule(l){6-9}
 & \multicolumn{2}{c}{\fast} & \multicolumn{2}{c}{\precise} & \multicolumn{2}{c}{\fast} & \multicolumn{2}{c@{}}{\precise} \\
\midrule
Time per frame, median (ms) & \best{81.7} &  & 116.4 &  & 187.8 &  & 278.0 &  \\
Time per frame, p95 (ms) & \best{251.1} &  & 417.4 &  & 677.3 &  & 1,030 &  \\
Peak GPU memory (GB) & \best{1.44} &  & \best{1.44} &  & 1.55 &  & 1.55 &  \\
\midrule
\multicolumn{9}{@{}l}{\textit{Recall at 5\,cm, clean frames}} \\
\quad ScanNet-10 & 75.7 & \,{\tiny$\pm$6.5} & 79.2 & \,{\tiny$\pm$5.9} & 76.6 & \,{\tiny$\pm$6.5} & \best{79.7} & \,{\tiny$\pm$5.6} \\
\quad 7-Scenes & 93.0 & \,{\tiny$\pm$2.5} & 93.1 & \,{\tiny$\pm$2.7} & 93.6 & \,{\tiny$\pm$2.3} & \best{93.7} & \,{\tiny$\pm$2.7} \\
\quad 12-Scenes & 97.9 & \,{\tiny$\pm$3.1} & 98.6 & \,{\tiny$\pm$2.0} & 98.0 & \,{\tiny$\pm$2.9} & \best{98.7} & \,{\tiny$\pm$1.9} \\
\quad ScanNet-10-S2 & 73.3 & \,{\tiny$\pm$7.4} & 75.9 & \,{\tiny$\pm$7.3} & 73.5 & \,{\tiny$\pm$7.4} & \best{76.2} & \,{\tiny$\pm$7.4} \\
\midrule
\multicolumn{9}{@{}l}{\textit{Recall at 0.25\,m, mean over 15 color shifts}} \\
\quad ScanNet-10 & 64.0 & \,{\tiny$\pm$4.7} & 65.8 & \,{\tiny$\pm$4.4} & 74.6 & \,{\tiny$\pm$4.7} & \best{76.6} & \,{\tiny$\pm$4.2} \\
\quad 7-Scenes & 71.4 & \,{\tiny$\pm$5.8} & 72.8 & \,{\tiny$\pm$5.6} & 83.1 & \,{\tiny$\pm$4.4} & \best{84.3} & \,{\tiny$\pm$4.0} \\
\bottomrule
\end{tabular}

\caption{\textbf{Optional refinement and color adjustment trade time for accuracy or tolerance to color shifts.}
Recall (\%) averages three training seeds, with interval half-widths over walks.
Bold marks the best value per row.
Time and memory follow \cref{tab:main}.
\tabpointsdash}
\label{tab:points}
\end{table}

To separate search scale from changes in scene geometry, we place rooms apart on a grid with pitch \pnum{prose.rooms.pitch_m}\,m.
They share no walls.
The ten-room and seven-room point clouds join ScanNet-10 and 7-Scenes respectively.
The largest joins all \pnum{building.rooms} rooms from which ScanNet-96 is drawn, totaling \pnum{building.points_m} million points.
Retrieval references use the image descriptor DINOv2~\cite{oquab2024dinov2} to rank rooms against \pnum{prose.rooms.retrieval_views} rendered views per room.
\emph{Retrieve-3 + BUFFER-X} registers in the top three rooms.
\emph{Retrieve-1 + \fast} runs \ours in the top room.

On the ten-room point cloud, \fast reaches \pnum{rooms.scannet.fast.recall5}\% within 5\,cm in \pnum{rooms.scannet.fast.ms}\,ms per frame (\cref{fig:teaser,fig:rooms}a).
Its recall is equivalent to BUFFER-X given the true room.
Retrieval includes the true room in its top three for only \pnum{rooms.scannet.retrieval.top3}\% of frames.
Retrieve-3 + BUFFER-X therefore trails \fast by \pnum{rooms.scannet.fast_minus_rr3} points and takes \pnum{rooms.scannet.rr3_over_fast} times as long.

Searching the \pnum{building.rooms}-room point cloud gives \pnum{building.fast.r5_all}\% recall, only \pnum{building.fast.loss} points below a paired run given the true room.
That reference is separate from the room-given run in \cref{tab:unseen}.
Keeping several candidates is important: with one candidate, recall falls to \pnum{building.one_candidate.r5_all}\%.
Retrieve-1 + \fast reaches only \pnum{building.rr1.r5}\%, losing \pnum{building.fast_minus_rr1} points relative to searching all rooms.
The full search takes \pnum{building.fast.ms_all}\,ms and \pnum{building.peak_gb}\,GB per frame (\cref{fig:rooms}b).
RGB GeoTransformer and ColorPCR exhaust the GPU from \pnum{building.color_oom_rooms} rooms onward.

\subsection{A hospital emergency department}
\label{sec:hospital}
\begin{figure}[t]
\centering
\includegraphics[width=0.70\linewidth]{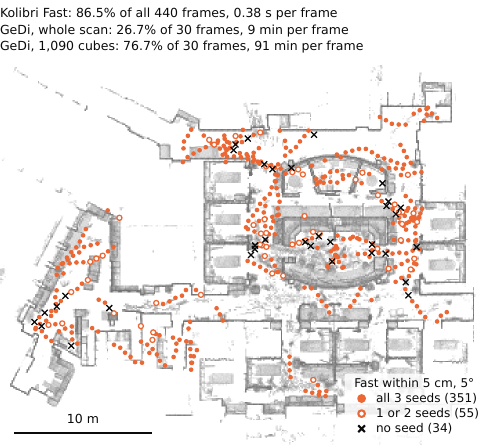}
\caption{\textbf{\ours localizes frames throughout the emergency department, with failures concentrated in a few places.}
The plotted positions belong to \pnum{hosp.frames} views rendered from the scan along \pnum{hosp.walks} walks, with simulated depth noise and colors that agree with the point cloud.
The floor plan comes from the scan geometry.
Markers count the training seeds for which \fast succeeds within 5\,cm and 5\textdegree.
GeDi uses the first frame of each walk.}
\label{fig:hospital}
\end{figure}

\begin{table}[t]
\centering
\scriptsize
\setlength{\tabcolsep}{3pt}
\begin{tabular}{@{}llrrrr@{}}
\toprule
Method & Search & Frames & 5\,cm & 0.5\,m & s/frame \\
\midrule
\oursrow \ours \fast & whole scan & 440 & 86.5\,{\tiny$\pm$0.7} & 90.6\,{\tiny$\pm$0.5} &\oursend \best{0.38} \\
\oursrow  &  & 30 & 84.4 & 90.0 &\oursend  \\
\oursrow \ours \precise & whole scan & 440 & 86.4\,{\tiny$\pm$0.5} & 91.1\,{\tiny$\pm$0.5} &\oursend 0.62 \\
\oursrow  &  & 30 & \best{86.7} & \best{91.1} &\oursend  \\
\midrule
GeDi & whole scan & 30 & 26.7 & 30.0 & 531 \\
GeDi & 350k keypoints, 15\,cm & 30 & 43.3 & 46.7 & 121 \\
GeDi & 1,090 cubes of 3\,m & 30 & 76.7 & 86.7 & 5,485 \\
BUFFER-X & 1,090 cubes of 3\,m & 30 & 60.0 & 63.3 & 932 \\
\bottomrule
\end{tabular}

\caption{\textbf{In a hospital emergency department, \ours localizes a frame in a fraction of a second, where GeDi needs \pnum{hosp.gedi_tiles.min} minutes on cubes of the scan for a similar recall.}
Views are rendered from the scan with simulated depth noise and matching colors.
Recall (\%) uses 5\,cm and 5\textdegree{} or 0.5\,m and 5\textdegree, without a room label.
Baselines use shared refinement on \pnum{hosp.sub_frames} frames, one per walk.
For \ours on all frames, $\pm$ gives the standard deviation over three training seeds.
Time is median seconds per frame on an RTX 2080 Ti.
The paired comparison between \fast and GeDi on cubes is inconclusive.}
\label{tab:hospital}
\end{table}

The emergency department returns to our motivating setting: many connected rooms captured on foot with a handheld laser scanner and colored by its camera.
The colored point cloud spans \pnum{hosp.extent_x}\,m by \pnum{hosp.extent_y}\,m in plan and played no part in development or tuning.
There is no RGB-D recording.
The test therefore contains \pnum{hosp.frames} RGB-D frames rendered from the scan along \pnum{hosp.walks} walks, with simulated depth noise.
Their colors agree with the point cloud by construction, which favors color-based methods, and their reference poses are exact.
\ours searches the whole scan without a room label.
Baselines take minutes per frame, so they use \pnum{hosp.sub_frames} frames, one from the start of each walk.

\fast localizes \pnum{hosp.fast.r5}\% of all rendered frames within 5\,cm and 5\textdegree, in \pnum{hosp.fast.ms}\,ms per frame on an RTX 2080 Ti (\cref{fig:hospital,tab:hospital}).
GeDi on the whole point cloud reaches \pnum{hosp.gedi_whole.r5}\% on the baseline subset.
Its random sample consensus (RANSAC) estimator fits poses from sampled matches, but ends with fewer than three agreeing matches on most frames.
Searching \pnum{hosp.tiles} overlapping \pnum{hosp.tile_m}\,m cubes raises GeDi's recall to \pnum{hosp.gedi_tiles.r5}\%.
The cost is \pnum{hosp.gedi_tiles.min} minutes per frame.
On these same frames, the difference between \fast and GeDi on cubes is inconclusive: \pnum{hosp.fast_minus_gedi_tiles} points [\pnum{hosp.fast_minus_gedi_tiles.lo}, \pnum{hosp.fast_minus_gedi_tiles.hi}].
BUFFER-X on the cubes reaches \pnum{hosp.bufferx_tiles.r5}\%.

A sparser point cloud reduces this cost but does not close the gap.
\ours represents the scan with \pnum{hosp.cells} cells for place search.
Given a keypoint per cell, \pnum{hosp.gedi_sparse.keypoints} in all, GeDi reaches \pnum{hosp.gedi_sparse.r5}\% in \pnum{hosp.gedi_sparse.min} minutes per frame with its best of \pnum{hosp.gedi_sparse.n_rd} RANSAC inlier distances.
This improves both recall and time over GeDi on the full point cloud, but still takes \pnum{hosp.gedi_sparse_over_fast} times as long as \fast for half its recall.

\subsection{Operating points and dependence on color}
\label{sec:price}
The second refinement stage in \precise adds \pnum{tab_points.precise_extra_ms}\,ms per frame (\cref{tab:points}) and gains \pnum{campaign_a.precise_minus_fast} points at 5\,cm on ScanNet-96.
The \colorpass takes \pnum{tab_points.fast_color.time_ratio} times as long as \fast.
Across the tested color changes, including brightness shifts and conversion to gray, it gains an average \pnum{tab_points.fast_color.color_gain_scannet} points at 0.25\,m and 2\textdegree{} on ScanNet-10 and \pnum{tab_points.fast_color.color_gain_sevenscenes} on 7-Scenes.

These options do not remove the dependence on color.
At the same threshold, RGB GeoTransformer leads \ours by \pnum{robust.color.geotr_mean} points averaged over the color shifts, and by \pnum{robust.color.geotr_gray} on gray frames.
The same limitation appears across cameras.
Bonn~\cite{palazzolo2019refusion} pairs real ASUS Xtion frames with a terrestrial laser scan colored by the scanner.
\fast reaches \pnum{bonn.fast.r50}\% within 0.5\,m, or \pnum{bonn.fast_color.r50}\% with the \colorpass, against \pnum{bonn.gedi_tiles.r50}\% for GeDi on cubes.
GeDi takes \pnum{bonn.gedi_tiles.min} minutes per frame.
Changes between recording sessions affect \fast and the six strongest refined baselines alike: they lose \pnum{campaign_a.rescan.cost_min} to \pnum{campaign_a.rescan.cost_max} points at 5\,cm on ScanNet-96-S2 (\cref{tab:unseen}).

\section{Conclusion}
\label{sec:conclusion}
\ours uses color to find candidate places and geometric refinement to set the pose.
With matched refinement, a small network reaches the accuracy of point cloud registration methods in single rooms at a fraction of their cost.
The saved time makes it practical to search a point cloud spanning \pnum{building.rooms} rooms or a hospital emergency department without a room label.
Agreement in color remains essential, and point cloud registration methods retain the lead when it breaks.
Extending search to whole buildings calls for a hierarchy that reduces work without discarding the correct place before refinement.

\label{page:main_end}
\iftoggle{cvprfinal}{%
\section*{Author Contributions}
Contributions by role, following the CRediT taxonomy:
\textbf{Bartu Tuncay}: Conceptualization, Methodology, Software, Data curation, Formal analysis, Investigation, Validation, Visualization, Writing (original draft), Writing (review and editing).
\textbf{Leonel Aguilar}: Conceptualization, Methodology, Software, Formal analysis, Investigation, Validation, Visualization, Writing (original draft), Writing (review and editing), Supervision, Project administration.
\textbf{Christoph H\"olscher}: Supervision, Resources, Funding acquisition, Writing (review and editing).
}{}

\section*{LLM Usage Statement}
OpenAI Codex and Anthropic Claude helped edit the paper and supplement, check
reported values. Claude Code was used for programming the computational experiments.
{
    \small
    \bibliographystyle{ieeenat_fullname}
    \bibliography{refs}
}

\clearpage
\appendix
\setcounter{section}{0}\setcounter{table}{0}\setcounter{figure}{0}
\renewcommand{\thesection}{S\arabic{section}}
\renewcommand{\thetable}{S\arabic{table}}
\renewcommand{\thefigure}{S\arabic{figure}}
\section*{Supplementary Material}

This supplement provides the full comparison rows, protocols and controls behind the main paper.
The operating points \fast, \precise{} and the \colorpass are defined in Sec.~\ref{sec:switches}.
Statistical terms, including equivalent, lead and inconclusive, follow Sec.~\ref{sec:setup}.

\section{What color contributes}
\label{sec:supp_color}

\paragraph{Human scene recognition.}
People recognize scene categories from brief glimpses~\cite{greene2009briefest,schyns1994blobs}, using coarse visual structure before fine edges~\cite{schyns1994blobs}.
Color helps most at this coarse scale~\cite{oliva2000diagnostic,castelhano2008color}, especially when a scene category has characteristic colors~\cite{oliva2000diagnostic}.
Recognizing a place and orienting within it engage different brain systems~\cite{epstein2008parahippocampal,epstein2014neural}.
Facing direction is represented relative to the local environment~\cite{marchette2014anchoring}, and young children reorient using room shape even when a wall has a distinctive color~\cite{hermer1994geometric}.
This distinction has a parallel in \ours: color identifies candidate places, and geometry sets the pose.

\begin{figure}[t]
\centering
\includegraphics[width=\linewidth]{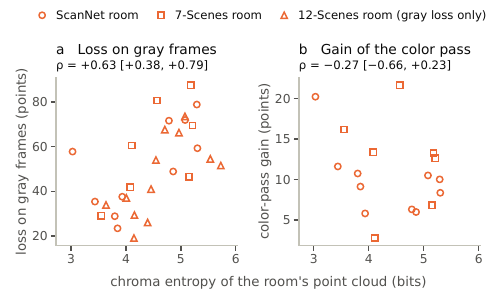}
\caption{\textbf{Rooms with more diverse colors lose more recall on gray frames.}
Each point represents a room.
Color diversity is the entropy of the chromatic coordinates $a^*$ and $b^*$ in the CIELAB color space.
(a)~The vertical axis gives the drop in \fast recall from unchanged to gray frames at 0.25\,m and 2\textdegree.
(b)~It gives the gain from the \colorpass, averaged over color shifts on ScanNet and 7-Scenes.}
\label{fig:diagnostic}
\end{figure}

\paragraph{Which rooms depend on color?}
Color diversity differs between rooms.
We measure it by the entropy of the point cloud cells' chromatic coordinates, which describe color independently of lightness.
Across \pnum{diag.rooms} rooms, greater diversity accompanies a larger drop in \fast recall when frames become gray (\cref{fig:diagnostic}a).
At 0.25\,m and 2\textdegree, Spearman's rank correlation $\rho$ is \pnum{diag.gray.rho}, with a room-bootstrap interval of $[\pnum{diag.gray.lo}, \pnum{diag.gray.hi}]$.
The relationship also holds within ScanNet and 7-Scenes, with correlations of \pnum{diag.within.scannet_gray} and \pnum{diag.within.sevenscenes_gray}.
For the gain from the \colorpass under color shifts, the corresponding correlation is $\rho=\pnum{diag.gain.rho}$ $[\pnum{diag.gain.lo}, \pnum{diag.gain.hi}]$, with an interval that includes zero (\cref{fig:diagnostic}b).

\begin{figure}[t]
\centering
\includegraphics[width=\linewidth]{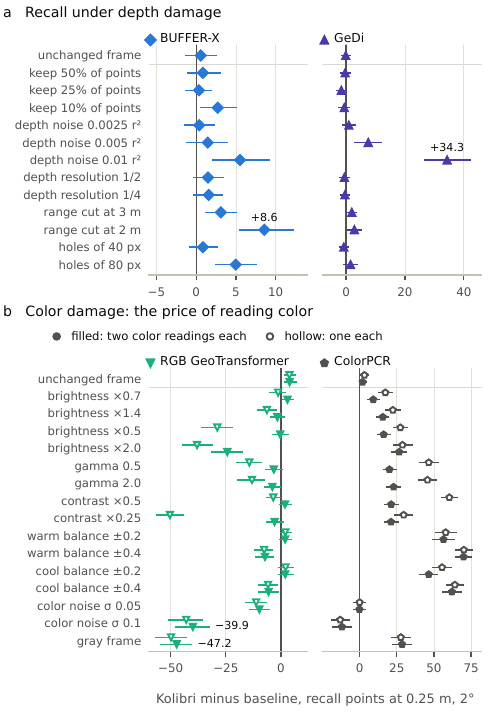}
\caption{\textbf{Damaged depth costs \ours no more than the baselines, and damaged color is its price.}
Each comparison gives \fast recall minus baseline recall at 0.25\,m and 2\textdegree, with intervals over walks.
The frame's depth changes in (a), and its color changes in (b).
The point cloud stays unchanged.
In (b), filled marks use two color readings selected by the same geometric fit score, corresponding to the \colorpass for \ours.
Hollow marks use one reading.}
\label{fig:robust_full}
\end{figure}

\paragraph{Changes to depth and color.}
\Cref{fig:robust_full} gives both panels of the damage tests from Sec.~\ref{sec:price}.
Across \pnum{robust.depth.levels} levels of point loss, depth noise, reduced resolution, range cuts and holes, \fast has no significant deficit against BUFFER-X or GeDi.
At 0.25\,m and 2\textdegree, its largest leads are \pnum{robust.depth.bufferx_max} and \pnum{robust.depth.gedi_max} points respectively (\cref{fig:robust_full}a).
Color changes expose the dependence on color (\cref{fig:robust_full}b).
On gray frames, RGB GeoTransformer leads \ours by \pnum{robust.color.geotr_gray} points at the same threshold.

\section{Several candidates against a lookalike place}
\label{sec:supp_lookalike}

\begin{figure*}[t]
\centering
\includegraphics[width=\linewidth]{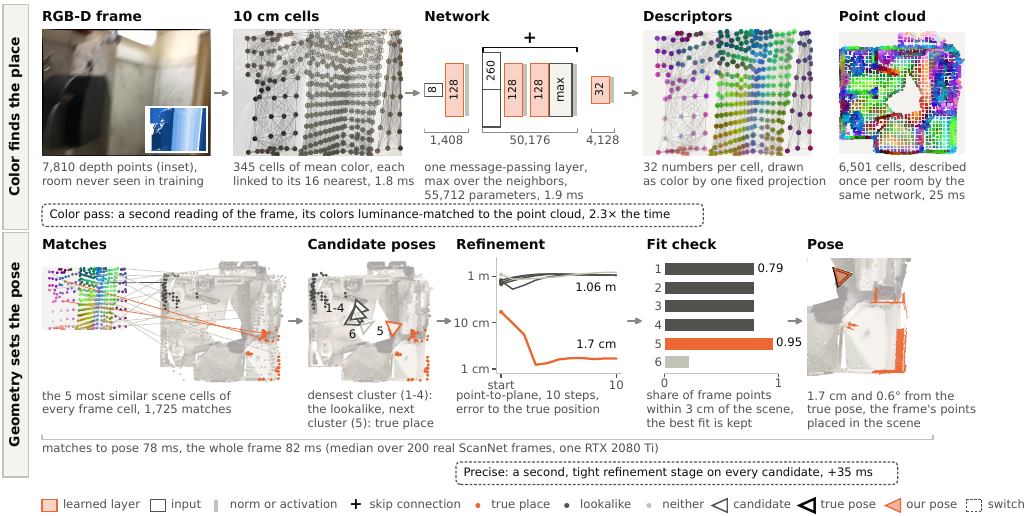}
\caption{\textbf{Several candidates recover the true place when the strongest match group points to a lookalike.}
The example is an unseen washroom with similar stalls, localized by \fast.
The top row shows the frame, its cell graph, the network blocks at their relative sizes, and descriptors colored by a shared projection.
The bottom row follows the matches through candidate poses, refinement, fit scoring and the final pose.
Dashed boxes mark the optional second refinement and color pass from Sec.~\ref{sec:switches}.}
\label{fig:method_example}
\end{figure*}

Repeated structures can give the strongest group of matches to the wrong place.
In the washroom example of \cref{fig:method_example}, that group points to a similar stall \pnum{method.lookalike_m}\,m away, rotated by \pnum{method.lookalike_deg}\textdegree.
Another group proposes the true place.
Refinement and the geometric fit score from Sec.~\ref{sec:pose} select it: \pnum{method.fit_true} of the frame points fall within \pnum{method.verify_cm}\,cm of the point cloud, compared with \pnum{method.fit_look} at the lookalike.
The final position error is \pnum{method.err_cm}\,cm.

\section{Full room-given comparisons}
\label{sec:supp_rows}

\paragraph{Development sets.}
\Cref{tab:main_full} extends Table~\ref{tab:main} with the second refinement stage for every baseline.
These rows compare against \precise, so the refinement is matched.

\begin{table*}[t]
\centering
\scriptsize
\setlength{\tabcolsep}{1.25pt}
\begin{tabular}{@{}l r c r r r r@{}l r@{}l r@{}l r@{}l r@{}l r@{}l r@{}l r@{}l@{}}
\toprule
 & & Test scenes & \multicolumn{3}{c}{Per frame} & \multicolumn{4}{c}{\begin{tabular}{@{}c@{}}ScanNet-10\\871 frames\end{tabular}} & \multicolumn{4}{c}{\begin{tabular}{@{}c@{}}7-Scenes\\1,700 frames\end{tabular}} & \multicolumn{4}{c}{\begin{tabular}{@{}c@{}}12-Scenes\\575 frames\end{tabular}} & \multicolumn{4}{c}{\begin{tabular}{@{}c@{}}ScanNet-10-S2\\683 frames\end{tabular}} \\
\cmidrule(lr){4-6}\cmidrule(lr){7-10}\cmidrule(lr){11-14}\cmidrule(lr){15-18}\cmidrule(l){19-22}
Method & Parameters & in training & median ms & p95 ms & GB & \multicolumn{2}{c}{5\,cm} & \multicolumn{2}{c}{0.5\,m} & \multicolumn{2}{c}{5\,cm} & \multicolumn{2}{c}{0.5\,m} & \multicolumn{2}{c}{5\,cm} & \multicolumn{2}{c}{0.5\,m} & \multicolumn{2}{c}{5\,cm} & \multicolumn{2}{c}{0.5\,m} \\
\midrule
\oursrow \ours \fast & 55.7k & none & \best{81.7} & \best{251.1} & \best{1.44} & 75.7{\tiny$\pm$6.5} &  & 94.9{\tiny$\pm$2.5} &  & 93.0{\tiny$\pm$2.5} &  & 98.5{\tiny$\pm$0.9} &  & 97.9{\tiny$\pm$3.1} &  & 99.7{\tiny$\pm$0.6} &  & 73.3{\tiny$\pm$7.4} &  & 95.3{\tiny$\pm$3.3} &\oursend  \\
\oursrow \ours \precise & 55.7k & none & 116.4 & 417.4 & \best{1.44} & 79.2{\tiny$\pm$5.9} &  & 95.1{\tiny$\pm$2.5} &  & \best{93.1}{\tiny$\pm$2.7} &  & \best{98.6}{\tiny$\pm$0.9} &  & 98.6{\tiny$\pm$2.0} &  & 99.7{\tiny$\pm$0.6} &  & 75.9{\tiny$\pm$7.3} &  & 95.4{\tiny$\pm$3.4} &\oursend  \\
\midrule
BUFFER-X~\cite{seo2025bufferx} & 0.91M & 7S, 12S & 1,416 & 1,922 & 6.21 & 66.4{\tiny$\pm$6.7} & $^{\blacktriangle}$ & 91.3{\tiny$\pm$3.8} & $^{\blacktriangle}$ & 72.7{\tiny$\pm$7.8} & $^{\blacktriangle}$ & 96.2{\tiny$\pm$2.5} & $^{\blacktriangle}$ & 82.1{\tiny$\pm$7.8} & $^{\blacktriangle}$ & 91.8{\tiny$\pm$5.0} & $^{\blacktriangle}$ & 66.9{\tiny$\pm$8.6} & $^{\blacktriangle}$ & 95.2{\tiny$\pm$2.9} & $^{\circ}$ \\
\quad + refinement & &  & 1,444 & 1,948 & 6.21 & 74.7{\tiny$\pm$6.9} & $^{\approx}$ & 92.9{\tiny$\pm$3.6} & $^{\circ}$ & 92.2{\tiny$\pm$3.6} & $^{\approx}$ & 97.5{\tiny$\pm$2.3} & $^{\circ}$ & 91.8{\tiny$\pm$4.7} & $^{\blacktriangle}$ & 93.4{\tiny$\pm$4.4} & $^{\blacktriangle}$ & 73.8{\tiny$\pm$7.3} & $^{\approx}$ & 97.1{\tiny$\pm$2.0} & $^{\circ}$ \\
\quad + second stage & &  & 1,480 & 1,997 & 6.21 & 77.4{\tiny$\pm$6.3} & $^{\blacktriangle}$ & 92.9{\tiny$\pm$3.6} & $^{\blacktriangle}$ & 92.1{\tiny$\pm$4.0} & $^{\circ}$ & 97.5{\tiny$\pm$2.3} & $^{\circ}$ & 92.5{\tiny$\pm$4.5} & $^{\blacktriangle}$ & 93.4{\tiny$\pm$4.6} & $^{\blacktriangle}$ & \best{76.6}{\tiny$\pm$6.8} & $^{\approx}$ & 97.1{\tiny$\pm$2.0} & $^{\circ}$ \\
\midrule
BUFFER-X-Lite~\cite{seo2025bufferx} & 0.91M & 7S, 12S & 1,280 & 1,896 & 6.21 & 63.9{\tiny$\pm$6.7} & $^{\blacktriangle}$ & 90.7{\tiny$\pm$4.2} & $^{\blacktriangle}$ & 66.6{\tiny$\pm$7.3} & $^{\blacktriangle}$ & 96.4{\tiny$\pm$2.5} & $^{\blacktriangle}$ & 81.6{\tiny$\pm$7.9} & $^{\blacktriangle}$ & 91.7{\tiny$\pm$5.6} & $^{\blacktriangle}$ & 62.1{\tiny$\pm$9.6} & $^{\blacktriangle}$ & 95.0{\tiny$\pm$2.9} & $^{\circ}$ \\
\quad + refinement & &  & 1,299 & 1,925 & 6.21 & 74.4{\tiny$\pm$6.8} & $^{\approx}$ & 92.9{\tiny$\pm$3.5} & $^{\circ}$ & 92.3{\tiny$\pm$3.5} & $^{\approx}$ & 97.5{\tiny$\pm$2.3} & $^{\circ}$ & 91.7{\tiny$\pm$4.7} & $^{\blacktriangle}$ & 93.4{\tiny$\pm$4.5} & $^{\blacktriangle}$ & 73.5{\tiny$\pm$7.0} & $^{\approx}$ & 96.8{\tiny$\pm$2.2} & $^{\circ}$ \\
\quad + second stage & &  & 1,317 & 1,949 & 6.21 & 77.4{\tiny$\pm$6.2} & $^{\circ}$ & 92.9{\tiny$\pm$3.7} & $^{\circ}$ & 92.3{\tiny$\pm$3.8} & $^{\approx}$ & 97.5{\tiny$\pm$2.2} & $^{\circ}$ & 92.3{\tiny$\pm$4.5} & $^{\blacktriangle}$ & 93.4{\tiny$\pm$4.5} & $^{\blacktriangle}$ & 76.3{\tiny$\pm$6.9} & $^{\approx}$ & 96.8{\tiny$\pm$2.1} & $^{\circ}$ \\
\midrule
GeDi~\cite{poiesi2022learning} & 3.11M & 7S, 12S & 24,149 & 43,869 & 5.27 & 78.8{\tiny$\pm$5.8} & $^{\circ}$ & 92.8{\tiny$\pm$3.5} & $^{\circ}$ & 70.8{\tiny$\pm$9.2} & $^{\blacktriangle}$ & 90.8{\tiny$\pm$3.9} & $^{\blacktriangle}$ & 98.8{\tiny$\pm$0.9} & $^{\circ}$ & 99.7{\tiny$\pm$0.4} & $^{\approx}$ & 74.4{\tiny$\pm$6.8} & $^{\circ}$ & 95.9{\tiny$\pm$2.5} & $^{\circ}$ \\
\quad + refinement & &  & 24,179 & 43,902 & 5.27 & 76.2{\tiny$\pm$7.1} & $^{\approx}$ & 95.1{\tiny$\pm$2.6} & $^{\approx}$ & 90.9{\tiny$\pm$3.8} & $^{\circ}$ & 95.8{\tiny$\pm$2.4} & $^{\blacktriangle}$ & 98.1{\tiny$\pm$2.9} & $^{\approx}$ & 99.7{\tiny$\pm$0.6} & $^{\approx}$ & 73.6{\tiny$\pm$7.1} & $^{\approx}$ & \best{97.5}{\tiny$\pm$1.7} & $^{\circ}$ \\
\quad + second stage & &  & 24,205 & 43,929 & 5.27 & \best{79.9}{\tiny$\pm$5.8} & $^{\approx}$ & \best{95.3}{\tiny$\pm$2.5} & $^{\approx}$ & 90.8{\tiny$\pm$4.1} & $^{\blacktriangle}$ & 95.8{\tiny$\pm$2.5} & $^{\blacktriangle}$ & 99.0{\tiny$\pm$1.6} & $^{\approx}$ & 99.7{\tiny$\pm$0.6} & $^{\approx}$ & \best{76.6}{\tiny$\pm$6.9} & $^{\approx}$ & \best{97.5}{\tiny$\pm$1.7} & $^{\circ}$ \\
\midrule
RGB GeoTransformer~\cite{qin2022geometric} & 9.83M & 7S, 12S & 2,501 & 6,338 & 2.70 & 76.3{\tiny$\pm$5.5} & $^{\circ}$ & 88.2{\tiny$\pm$4.4} & $^{\blacktriangle}$ & 66.1{\tiny$\pm$9.8} & $^{\blacktriangle}$ & 92.4{\tiny$\pm$3.2} & $^{\blacktriangle}$ & \best{99.1}{\tiny$\pm$0.8} & $^{\circ}$ & \best{99.8}{\tiny$\pm$0.2} & $^{\approx}$ & 72.6{\tiny$\pm$7.2} & $^{\circ}$ & 93.0{\tiny$\pm$3.4} & $^{\circ}$ \\
\quad + refinement & &  & 2,530 & 6,372 & 2.70 & 73.5{\tiny$\pm$7.4} & $^{\blacktriangle}$ & 89.4{\tiny$\pm$4.0} & $^{\blacktriangle}$ & 89.8{\tiny$\pm$4.3} & $^{\blacktriangle}$ & 94.9{\tiny$\pm$2.9} & $^{\blacktriangle}$ & 98.1{\tiny$\pm$2.9} & $^{\approx}$ & 99.7{\tiny$\pm$0.4} & $^{\approx}$ & 72.3{\tiny$\pm$7.1} & $^{\approx}$ & 94.3{\tiny$\pm$2.9} & $^{\circ}$ \\
\quad + second stage & &  & 2,560 & 6,395 & 2.70 & 76.1{\tiny$\pm$6.4} & $^{\blacktriangle}$ & 89.7{\tiny$\pm$4.2} & $^{\blacktriangle}$ & 89.6{\tiny$\pm$4.3} & $^{\blacktriangle}$ & 94.9{\tiny$\pm$2.7} & $^{\blacktriangle}$ & 98.8{\tiny$\pm$1.7} & $^{\approx}$ & 99.7{\tiny$\pm$0.4} & $^{\approx}$ & 75.0{\tiny$\pm$7.3} & $^{\circ}$ & 94.3{\tiny$\pm$2.9} & $^{\circ}$ \\
\midrule
ColorPCR~\cite{mu2024colorpcr} & 10.03M & 7S & 4,077 & 10,878 & 3.90 & 66.4{\tiny$\pm$6.8} & $^{\blacktriangle}$ & 89.1{\tiny$\pm$4.5} & $^{\blacktriangle}$ & 63.5{\tiny$\pm$8.8} & $^{\blacktriangle}$ & 89.1{\tiny$\pm$4.8} & $^{\blacktriangle}$ & 95.5{\tiny$\pm$2.3} & $^{\circ}$ & 98.4{\tiny$\pm$1.0} & $^{\approx\blacktriangle}$ & 58.6{\tiny$\pm$9.8} & $^{\blacktriangle}$ & 81.7{\tiny$\pm$7.6} & $^{\blacktriangle}$ \\
\quad + refinement & &  & 4,107 & 10,913 & 3.90 & 72.6{\tiny$\pm$7.0} & $^{\blacktriangle}$ & 91.4{\tiny$\pm$3.6} & $^{\blacktriangle}$ & 86.3{\tiny$\pm$5.0} & $^{\blacktriangle}$ & 92.4{\tiny$\pm$4.0} & $^{\blacktriangle}$ & 97.2{\tiny$\pm$2.9} & $^{\approx}$ & 98.8{\tiny$\pm$0.9} & $^{\approx}$ & 67.2{\tiny$\pm$8.7} & $^{\blacktriangle}$ & 86.1{\tiny$\pm$6.7} & $^{\blacktriangle}$ \\
\quad + second stage & &  & 4,131 & 10,943 & 3.90 & 76.2{\tiny$\pm$6.0} & $^{\blacktriangle}$ & 91.8{\tiny$\pm$3.6} & $^{\blacktriangle}$ & 86.1{\tiny$\pm$5.1} & $^{\blacktriangle}$ & 92.5{\tiny$\pm$4.0} & $^{\blacktriangle}$ & 97.9{\tiny$\pm$1.9} & $^{\approx}$ & 98.8{\tiny$\pm$0.9} & $^{\approx}$ & 69.4{\tiny$\pm$8.3} & $^{\blacktriangle}$ & 86.2{\tiny$\pm$6.5} & $^{\blacktriangle}$ \\
\midrule
RoITr~\cite{yu2023rotation} & 10.10M & 7S, 12S & 6,402 & 22,324 & 2.07 & 60.2{\tiny$\pm$6.9} & $^{\blacktriangle}$ & 86.3{\tiny$\pm$4.5} & $^{\blacktriangle}$ & 55.8{\tiny$\pm$13.3} & $^{\blacktriangle}$ & 81.2{\tiny$\pm$10.6} & $^{\blacktriangle}$ & 77.2{\tiny$\pm$8.9} & $^{\blacktriangle}$ & 92.5{\tiny$\pm$5.8} & $^{\blacktriangle}$ & 61.5{\tiny$\pm$6.9} & $^{\blacktriangle}$ & 90.6{\tiny$\pm$4.0} & $^{\circ}$ \\
\quad + refinement & &  & 6,429 & 22,353 & 2.07 & 72.1{\tiny$\pm$7.3} & $^{\blacktriangle}$ & 89.2{\tiny$\pm$4.0} & $^{\blacktriangle}$ & 81.4{\tiny$\pm$9.1} & $^{\blacktriangle}$ & 86.1{\tiny$\pm$9.0} & $^{\blacktriangle}$ & 92.9{\tiny$\pm$6.8} & $^{\blacktriangle}$ & 94.4{\tiny$\pm$4.9} & $^{\blacktriangle}$ & 71.4{\tiny$\pm$7.3} & $^{\circ}$ & 93.6{\tiny$\pm$3.5} & $^{\circ}$ \\
\quad + second stage & &  & 6,450 & 22,389 & 2.07 & 75.0{\tiny$\pm$6.7} & $^{\blacktriangle}$ & 89.7{\tiny$\pm$4.1} & $^{\blacktriangle}$ & 81.6{\tiny$\pm$9.2} & $^{\blacktriangle}$ & 86.2{\tiny$\pm$8.4} & $^{\blacktriangle}$ & 93.7{\tiny$\pm$5.6} & $^{\blacktriangle}$ & 94.8{\tiny$\pm$4.8} & $^{\blacktriangle}$ & 73.9{\tiny$\pm$7.0} & $^{\circ}$ & 93.6{\tiny$\pm$3.5} & $^{\circ}$ \\
\midrule
PREDATOR~\cite{huang2021predator} & 7.43M & 7S, 12S & 728.1 & 1,287 & 3.88 & 13.4{\tiny$\pm$4.3} & $^{\blacktriangle}$ & 38.2{\tiny$\pm$6.9} & $^{\blacktriangle}$ & 14.2{\tiny$\pm$8.1} & $^{\blacktriangle}$ & 33.7{\tiny$\pm$12.2} & $^{\blacktriangle}$ & 30.4{\tiny$\pm$8.8} & $^{\blacktriangle}$ & 52.9{\tiny$\pm$10.6} & $^{\blacktriangle}$ & 20.6{\tiny$\pm$6.6} & $^{\blacktriangle}$ & 46.9{\tiny$\pm$7.7} & $^{\blacktriangle}$ \\
\quad + refinement & &  & 754.0 & 1,316 & 3.88 & 47.4{\tiny$\pm$7.6} & $^{\blacktriangle}$ & 53.4{\tiny$\pm$7.0} & $^{\blacktriangle}$ & 47.0{\tiny$\pm$13.3} & $^{\blacktriangle}$ & 51.4{\tiny$\pm$12.8} & $^{\blacktriangle}$ & 72.5{\tiny$\pm$8.5} & $^{\blacktriangle}$ & 74.4{\tiny$\pm$9.0} & $^{\blacktriangle}$ & 53.1{\tiny$\pm$7.9} & $^{\blacktriangle}$ & 67.6{\tiny$\pm$6.6} & $^{\blacktriangle}$ \\
\quad + second stage & &  & 783.7 & 1,347 & 3.88 & 49.1{\tiny$\pm$7.1} & $^{\blacktriangle}$ & 53.4{\tiny$\pm$6.8} & $^{\blacktriangle}$ & 48.6{\tiny$\pm$13.3} & $^{\blacktriangle}$ & 51.9{\tiny$\pm$12.8} & $^{\blacktriangle}$ & 73.4{\tiny$\pm$9.0} & $^{\blacktriangle}$ & 74.6{\tiny$\pm$9.2} & $^{\blacktriangle}$ & 55.3{\tiny$\pm$7.2} & $^{\blacktriangle}$ & 68.7{\tiny$\pm$6.6} & $^{\blacktriangle}$ \\
\midrule
FPFH + TEASER++~\cite{rusu2009fast,yang2020teaser} & none & none & 215.0 & 648.5 & -- & 11.0{\tiny$\pm$3.9} & $^{\blacktriangle}$ & 24.9{\tiny$\pm$6.7} & $^{\blacktriangle}$ & 2.6{\tiny$\pm$1.5} & $^{\blacktriangle}$ & 10.8{\tiny$\pm$4.5} & $^{\blacktriangle}$ & 8.0{\tiny$\pm$2.9} & $^{\blacktriangle}$ & 16.3{\tiny$\pm$5.0} & $^{\blacktriangle}$ & 9.5{\tiny$\pm$3.4} & $^{\blacktriangle}$ & 29.6{\tiny$\pm$8.0} & $^{\blacktriangle}$ \\
\quad + refinement & &  & 236.9 & 701.0 & -- & 30.7{\tiny$\pm$7.4} & $^{\blacktriangle}$ & 35.2{\tiny$\pm$7.2} & $^{\blacktriangle}$ & 19.5{\tiny$\pm$6.3} & $^{\blacktriangle}$ & 20.6{\tiny$\pm$6.5} & $^{\blacktriangle}$ & 31.3{\tiny$\pm$7.3} & $^{\blacktriangle}$ & 32.3{\tiny$\pm$6.9} & $^{\blacktriangle}$ & 33.7{\tiny$\pm$8.0} & $^{\blacktriangle}$ & 45.7{\tiny$\pm$9.1} & $^{\blacktriangle}$ \\
\quad + second stage & &  & 265.2 & 740.9 & -- & 32.7{\tiny$\pm$7.2} & $^{\blacktriangle}$ & 35.5{\tiny$\pm$7.0} & $^{\blacktriangle}$ & 19.9{\tiny$\pm$6.8} & $^{\blacktriangle}$ & 21.1{\tiny$\pm$6.8} & $^{\blacktriangle}$ & 32.2{\tiny$\pm$7.2} & $^{\blacktriangle}$ & 33.0{\tiny$\pm$7.0} & $^{\blacktriangle}$ & 35.7{\tiny$\pm$8.3} & $^{\blacktriangle}$ & 46.3{\tiny$\pm$8.7} & $^{\blacktriangle}$ \\
\midrule
\multicolumn{3}{@{}l}{\textit{Trained on each test scene}} & \multicolumn{3}{c}{\textit{training, per frame}} & \multicolumn{16}{l}{} \\
ACE~\cite{brachmann2023accelerated} (RGB) & 2.1M + 5.6M & our run & \multicolumn{3}{c}{258\,s (291\,s$^\dagger$), 29\,ms} & -- &  & -- &  & 76.8{\tiny$\pm$8.1} &  & 99.4{\tiny$\pm$0.6} &  & 99.5{\tiny$\pm$0.6} &  & 100.0{\tiny$\pm$0.0} &  & -- &  & -- &  \\
DSAC*~\cite{brachmann2021visual} (RGB) & 28\,MB$^\dagger$ & from~\cite{brachmann2023accelerated,brachmann2021visual} & \multicolumn{3}{c}{15\,h$^\dagger$, 75\,ms$^\dagger$} & -- &  & -- &  & 84.0 & $^\dagger$ & -- &  & 99.2 & $^\dagger$ & -- &  & -- &  & -- &  \\
DSAC*~\cite{brachmann2021visual} (RGB-D) & 28\,MB$^\dagger$ & from~\cite{brachmann2021limits,brachmann2021visual} & \multicolumn{3}{c}{2.5\,d$^\dagger$} & -- &  & -- &  & 94.7 & $^\dagger$ & -- &  & -- &  & -- &  & -- &  & -- &  \\
\bottomrule
\end{tabular}

\caption{\textbf{Room-given development sets with every refinement stage.}
Recall (\%) uses 5\,cm and 5\textdegree{} or 0.5\,m and 5\textdegree.
Gray rows are \ours, and bold marks the best value per column among methods that do not train on the test scenes.
\ours averages three training seeds, while each baseline uses one run.
The $\pm$ entries give half-widths of intervals over walks.
Native denotes released output, including the refinement supplied by BUFFER-X and BUFFER-X-Lite.
The next rows add our point-to-plane refinement and then the second stage.
Marks compare \fast with native and refined rows, and \precise with second-stage rows: $\approx$ equivalent within $\pm$\pnum{tab_main.margin} points, $\blacktriangle$ \ours leads, $\circ$ inconclusive, $\approx\blacktriangle$ both.
Time and memory are medians over \pnum{tab_main.timing_frames} ScanNet-10 frames on an RTX 2080 Ti (for ACE, over its scenes).
FPFH + TEASER++ uses the CPU.
The third column identifies test scenes present in training data, or the source of a row in the last block.
Methods in the last block train on each test scene, and $^\dagger$ marks values from the cited papers, on their full test splits and the GPUs they name.}
\label{tab:main_full}
\end{table*}

\paragraph{Held-out test.}
\Cref{tab:final} gives the full results on \pnum{final.rooms} rooms evaluated once after the method was frozen.
The set contains \pnum{final.real.n} sensor frames and \pnum{final.rendered.n} rendered views.
With matched refinement, \fast and \precise are equivalent to GeDi on sensor frames at 5\,cm and 5\textdegree{} and at 0.25\,m and 2\textdegree.
Comparisons with the other baselines favor \ours or are inconclusive.
GeDi's native output outperforms \fast at 5\,cm and 5\textdegree, but this lead disappears when both use the shared refinement.
On rendered views, matched refinement makes \ours equivalent to ColorPCR and ahead of the remaining baselines.

\begin{table*}[t]
\centering
\scriptsize
\setlength{\tabcolsep}{2.2pt}
\begin{tabular}{@{}l r@{}l r@{}l r@{}l r@{}l r@{}l r@{}l@{}}
\toprule
 & \multicolumn{6}{c}{Sensor frames (679)} & \multicolumn{6}{c}{Rendered views (2,000)} \\
\cmidrule(lr){2-7}\cmidrule(l){8-13}
Method & \multicolumn{2}{c}{5\,cm} & \multicolumn{2}{c}{0.25\,m} & \multicolumn{2}{c}{0.5\,m} & \multicolumn{2}{c}{5\,cm} & \multicolumn{2}{c}{0.25\,m} & \multicolumn{2}{c}{0.5\,m} \\
\midrule
\oursrow \ours \fast & 68.7 &  & 84.9 &  & 92.9 &  & 94.6 &  & 94.7 &  & 94.9 &\oursend  \\
\oursrow \ours \precise & 73.1 &  & 87.1 &  & 93.3 &  & \best{94.8} &  & \best{94.9} &  & \best{95.0} &\oursend  \\
\oursrow \ours \fast + \colorpass & 69.1 &  & 86.2 &  & 93.9 &  & 94.5 &  & 94.7 &  & 94.8 &\oursend  \\
\oursrow \ours \precise + \colorpass & \best{73.4} &  & \best{88.0} &  & \best{94.4} &  & \best{94.8} &  & \best{94.9} &  & 94.9 &\oursend  \\
\midrule
BUFFER-X~\cite{seo2025bufferx} & 58.3 & $^{\blacktriangle}$ & 72.9 & $^{\blacktriangle}$ & 86.6 & $^{\blacktriangle}$ & 86.2 & $^{\blacktriangle}$ & 85.7 & $^{\blacktriangle}$ & 89.2 & $^{\blacktriangle}$ \\
\quad + refinement & 66.0 & $^{\circ}$ & 81.7 & $^{\circ}$ & 88.1 & $^{\circ}$ & 90.6 & $^{\blacktriangle}$ & 90.7 & $^{\blacktriangle}$ & 90.8 & $^{\blacktriangle}$ \\
\quad + second stage & 69.2 & $^{\circ}$ & 83.2 & $^{\circ}$ & 88.4 & $^{\circ}$ & 90.8 & $^{\blacktriangle}$ & 90.8 & $^{\blacktriangle}$ & 90.8 & $^{\blacktriangle}$ \\
\midrule
BUFFER-X-Lite~\cite{seo2025bufferx} & 54.3 & $^{\blacktriangle}$ & 68.8 & $^{\blacktriangle}$ & 86.2 & $^{\blacktriangle}$ & 83.2 & $^{\blacktriangle}$ & 85.3 & $^{\blacktriangle}$ & 89.4 & $^{\blacktriangle}$ \\
\quad + refinement & 66.4 & $^{\circ}$ & 81.9 & $^{\circ}$ & 87.8 & $^{\circ}$ & 90.8 & $^{\blacktriangle}$ & 90.9 & $^{\blacktriangle}$ & 91.0 & $^{\blacktriangle}$ \\
\quad + second stage & 69.7 & $^{\circ}$ & 83.4 & $^{\circ}$ & 88.1 & $^{\circ}$ & 91.0 & $^{\blacktriangle}$ & 91.0 & $^{\blacktriangle}$ & 91.0 & $^{\blacktriangle}$ \\
\midrule
GeDi~\cite{poiesi2022learning} & 73.0 & $^{\blacktriangledown}$ & 82.2 & $^{\circ}$ & 89.8 & $^{\circ}$ & 88.1 & $^{\blacktriangle}$ & 88.1 & $^{\blacktriangle}$ & 89.6 & $^{\blacktriangle}$ \\
\quad + refinement & 68.6 & $^{\approx}$ & 84.8 & $^{\approx}$ & 91.3 & $^{\circ}$ & 90.9 & $^{\blacktriangle}$ & 91.0 & $^{\blacktriangle}$ & 91.0 & $^{\blacktriangle}$ \\
\quad + second stage & 72.6 & $^{\approx}$ & 86.7 & $^{\approx}$ & 91.9 & $^{\circ}$ & 91.0 & $^{\blacktriangle}$ & 91.0 & $^{\blacktriangle}$ & 91.1 & $^{\blacktriangle}$ \\
\midrule
RGB GeoTransformer~\cite{qin2022geometric} & 64.4 & $^{\circ}$ & 71.0 & $^{\blacktriangle}$ & 78.1 & $^{\blacktriangle}$ & 86.2 & $^{\blacktriangle}$ & 85.5 & $^{\blacktriangle}$ & 87.4 & $^{\blacktriangle}$ \\
\quad + refinement & 62.7 & $^{\blacktriangle}$ & 74.7 & $^{\blacktriangle}$ & 79.5 & $^{\blacktriangle}$ & 87.8 & $^{\blacktriangle}$ & 87.8 & $^{\blacktriangle}$ & 87.8 & $^{\blacktriangle}$ \\
\quad + second stage & 65.4 & $^{\blacktriangle}$ & 75.7 & $^{\blacktriangle}$ & 80.0 & $^{\blacktriangle}$ & 87.8 & $^{\blacktriangle}$ & 87.8 & $^{\blacktriangle}$ & 87.8 & $^{\blacktriangle}$ \\
\midrule
ColorPCR~\cite{mu2024colorpcr} & 63.8 & $^{\circ}$ & 74.1 & $^{\blacktriangle}$ & 84.5 & $^{\blacktriangle}$ & 92.0 & $^{\blacktriangle}$ & 91.4 & $^{\blacktriangle}$ & 93.6 & $^{\approx\blacktriangle}$ \\
\quad + refinement & 65.7 & $^{\blacktriangle}$ & 80.9 & $^{\blacktriangle}$ & 87.5 & $^{\blacktriangle}$ & 94.2 & $^{\approx}$ & 94.2 & $^{\approx}$ & 94.3 & $^{\approx}$ \\
\quad + second stage & 69.4 & $^{\blacktriangle}$ & 82.2 & $^{\blacktriangle}$ & 87.5 & $^{\blacktriangle}$ & 94.2 & $^{\approx}$ & 94.3 & $^{\approx}$ & 94.4 & $^{\approx}$ \\
\midrule
RoITr~\cite{yu2023rotation} & 49.5 & $^{\blacktriangle}$ & 58.9 & $^{\blacktriangle}$ & 71.9 & $^{\blacktriangle}$ & 44.8 & $^{\blacktriangle}$ & 46.5 & $^{\blacktriangle}$ & 59.9 & $^{\blacktriangle}$ \\
\quad + refinement & 60.1 & $^{\blacktriangle}$ & 70.8 & $^{\blacktriangle}$ & 73.9 & $^{\blacktriangle}$ & 64.5 & $^{\blacktriangle}$ & 64.7 & $^{\blacktriangle}$ & 65.0 & $^{\blacktriangle}$ \\
\quad + second stage & 62.6 & $^{\blacktriangle}$ & 71.6 & $^{\blacktriangle}$ & 74.2 & $^{\blacktriangle}$ & 64.8 & $^{\blacktriangle}$ & 64.9 & $^{\blacktriangle}$ & 65.1 & $^{\blacktriangle}$ \\
\midrule
PREDATOR~\cite{huang2021predator} & 15.8 & $^{\blacktriangle}$ & 16.5 & $^{\blacktriangle}$ & 33.1 & $^{\blacktriangle}$ & 25.3 & $^{\blacktriangle}$ & 27.2 & $^{\blacktriangle}$ & 45.7 & $^{\blacktriangle}$ \\
\quad + refinement & 40.2 & $^{\blacktriangle}$ & 45.2 & $^{\blacktriangle}$ & 47.6 & $^{\blacktriangle}$ & 55.2 & $^{\blacktriangle}$ & 55.9 & $^{\blacktriangle}$ & 56.6 & $^{\blacktriangle}$ \\
\quad + second stage & 42.0 & $^{\blacktriangle}$ & 46.2 & $^{\blacktriangle}$ & 48.0 & $^{\blacktriangle}$ & 55.8 & $^{\blacktriangle}$ & 56.2 & $^{\blacktriangle}$ & 56.7 & $^{\blacktriangle}$ \\
\midrule
FPFH + TEASER++~\cite{rusu2009fast,yang2020teaser} & 12.1 & $^{\blacktriangle}$ & 14.4 & $^{\blacktriangle}$ & 28.9 & $^{\blacktriangle}$ & 29.9 & $^{\blacktriangle}$ & 37.5 & $^{\blacktriangle}$ & 50.3 & $^{\blacktriangle}$ \\
\quad + refinement & 29.6 & $^{\blacktriangle}$ & 33.9 & $^{\blacktriangle}$ & 36.2 & $^{\blacktriangle}$ & 56.7 & $^{\blacktriangle}$ & 56.8 & $^{\blacktriangle}$ & 57.5 & $^{\blacktriangle}$ \\
\quad + second stage & 30.5 & $^{\blacktriangle}$ & 34.8 & $^{\blacktriangle}$ & 36.4 & $^{\blacktriangle}$ & 57.0 & $^{\blacktriangle}$ & 57.1 & $^{\blacktriangle}$ & 57.6 & $^{\blacktriangle}$ \\
\bottomrule
\end{tabular}

\caption{\textbf{Every row of the held-out test.}
Recall (\%) uses 5\,cm and 5\textdegree, 0.25\,m and 2\textdegree, or 0.5\,m and 5\textdegree.
Gray rows are \ours, and bold marks the best value per column.
A frame without a valid pose counts as a miss.
\ours averages three training seeds, while each baseline uses one run.
Native denotes released output, including the refinement supplied by BUFFER-X and BUFFER-X-Lite.
The next rows add our point-to-plane refinement and then the second stage.
Marks compare \fast with native and refined rows, and \precise with second-stage rows.
They follow \cref{tab:main_full}, with $\blacktriangledown$ indicating that \ours trails.}
\label{tab:final}
\end{table*}

\section{Search without a room label}
\label{sec:supp_scale}

The point cloud in Sec.~\ref{sec:rooms} joins \pnum{building.rooms} ScanNet rooms placed apart on a grid, totaling \pnum{building.points_m} million points.
Queries are all \pnum{building.frames} first-session frames of ScanNet-96.
\Cref{tab:supp_scale} compares search over all rooms with a paired run on the same frames given the true room.
This reference run is separate from the room-given results in Table~\ref{tab:unseen}.

\begin{table}[t]
\centering
\scriptsize\setlength{\tabcolsep}{3pt}
\begin{tabular}{@{}lrrrrr@{}}
\toprule
 & \multicolumn{2}{c}{Recall (\%)} & & \multicolumn{2}{c}{Median ms} \\
\cmidrule(lr){2-3}\cmidrule(l){5-6}
 & Own room & All rooms & Loss (points) & Own room & All rooms \\
\midrule
\fast & \pnum{building.fast.r5_one} & \pnum{building.fast.r5_all} & \pnum{building.fast.loss}\,[\pnum{building.fast.loss.lo}, \pnum{building.fast.loss.hi}] & \pnum{building.fast.ms_one} & \pnum{building.fast.ms_all} \\
\precise & \pnum{building.precise.r5_one} & \pnum{building.precise.r5_all} & \pnum{building.precise.loss}\,[\pnum{building.precise.loss.lo}, \pnum{building.precise.loss.hi}] & -- & \pnum{building.precise.ms_all} \\
\bottomrule
\end{tabular}
\caption{\textbf{Searching all rooms has a small recall cost.}
Recall uses 5\,cm and 5\textdegree.
The loss compares all-room search with a paired run given the true room, with an interval over walks.
Median times use the \pnum{building.timing_frames}-frame timing subset on an RTX 2080 Ti.}
\label{tab:supp_scale}
\end{table}

\paragraph{Candidates and room retrieval.}
Several candidates are important even with refinement fixed.
Keeping only one reduces all-room recall to \pnum{building.one_candidate.r5_all}\%.
The image descriptor DINOv2 ranks the true room first for \pnum{building.retrieval.top1}\% of frames and in the top three for \pnum{building.retrieval.top3}\%.
Running \fast in the top room gives \pnum{building.rr1.r5}\% recall, \pnum{building.fast_minus_rr1} points below searching all rooms, with interval $[\pnum{building.fast_minus_rr1.lo}, \pnum{building.fast_minus_rr1.hi}]$.
All-room search places \pnum{building.fast.right_room}\% of frames in the correct room.

\paragraph{Memory.}
Peak GPU memory is \pnum{building.peak_gb}\,GB for \ours on the full point cloud.
RGB GeoTransformer and ColorPCR exhaust the GPU from \pnum{building.color_oom_rooms} rooms onward.

\paragraph{Cell size.}
Both finer and coarser cells reduce recall on the ScanNet-10 ablation rooms.
Relative to the \pnum{method.cell_cm}\,cm default, cells of \pnum{cells.fine_cm}\,cm and \pnum{cells.coarse_cm}\,cm lose \pnum{cells.c05.loss_heldout} and \pnum{cells.c20.loss_heldout} points at 5\,cm and 5\textdegree.
Median times are \pnum{cells.c05.ms}, \pnum{cells.c10.ms} and \pnum{cells.c20.ms}\,ms for fine, default and coarse cells respectively.
Fine cells exhaust the RTX 2080 Ti on \pnum{cells.c05.oom_pct}\% of ScanNet-10 frames, so their recall is measured on an RTX 4090.

\section{The hospital emergency department}
\label{sec:supp_hospital}

\paragraph{Scan and frames.}
The colored point cloud contains \pnum{hosp.points_m} million points and spans \pnum{hosp.extent_x} by \pnum{hosp.extent_y}\,m in plan.
There is no RGB-D recording, so the test frames are rendered from the scan.
Each of the \pnum{hosp.walks} walks starts at a random walkable position and takes up to \pnum{hosp.max_steps} steps of \pnum{hosp.step_m}\,m.
The camera height ranges from \pnum{hosp.cam_h_lo_m} to \pnum{hosp.cam_h_hi_m}\,m, with a downward tilt of up to \pnum{hosp.pitch_max_deg}\textdegree.
Each point is rendered in its own color as a square of half-size \pnum{hosp.splat_half_cm}\,cm.
Simulated sensor noise perturbs depth, while colors agree with the point cloud by construction.
Frames require depth on at least \pnum{hosp.min_cover_pct}\% of pixels and a median depth of at least \pnum{hosp.min_median_m}\,m.
This leaves \pnum{hosp.frames} frames, shown in Fig.~\ref{fig:hospital}.
Baselines use the first frame of each walk.

\paragraph{GeDi's settings on the whole point cloud.}
GeDi's starting configuration describes every point and uses an inlier distance of \pnum{gedi.cal_cm}\,cm, calibrated on room development frames.
This distance determines which matches agree with a candidate pose in its random sample consensus (RANSAC) estimator.
On the full point cloud, most frames end with fewer than three agreeing matches.
\Cref{tab:hospital_grid} varies both the number of described points and the inlier distance.
Sparser descriptions reduce time, while wider inlier distances allow agreement among more of the remaining matches.
Across the \pnum{hosp.gedi_grid.runs} tested settings, the best whole-point-cloud recall at 5\,cm and 5\textdegree{} is \pnum{hosp.gedi_grid.best_r5}\%.
Searching separate \pnum{hosp.tile_m}\,m cubes raises recall to \pnum{hosp.gedi_tiles.r5}\%, at \pnum{hosp.gedi_tiles.min} minutes per frame.

\paragraph{The size of the searched region.}
\Cref{tab:hospital_crops} separates GeDi's descriptors from the size of its search.
Regions of the scan are cut around each of the \pnum{hosp.crop.frames} frames with the true pose, and GeDi runs with its calibrated settings and no refinement.
On a \pnum{hosp.tile_m}\,m cube around the points a frame sees, GeDi places \pnum{hosp.crop.cube3.r5}\% of the frames within 5\,cm and 5\textdegree.
Recall falls steadily as the region grows, to \pnum{hosp.crop.sq32.r5}\% on the largest square, and the share of frames for which GeDi returns a pose falls from \pnum{hosp.crop.valid.cube3}\% to \pnum{hosp.crop.valid.sq32}\%.
The descriptors work in the emergency department, and the whole-scan result is the cost of matching against a large point cloud.
The cube search recovers recall by running many small searches and keeping the cube with the most agreeing matches.

\begin{table}[t]
\centering
\scriptsize\setlength{\tabcolsep}{2.5pt}
\begin{tabular}{@{}lrrrrr@{}}
\toprule
Search & Map keypoints & Inlier dist. & 5\,cm & 0.5\,m & Time \\
\midrule
GeDi, whole scan & 100,000 & 2\,cm$^{\ast}$ & 10.0 & 10.0 & 0.7\,min \\
GeDi, whole scan & 100,000 & 10\,cm & 13.3 & 16.7 & 0.6\,min \\
GeDi, whole scan & 100,000 & 20\,cm & 16.7 & 20.0 & 0.6\,min \\
GeDi, whole scan & 350,000 & 2\,cm$^{\ast}$ & 13.3 & 16.7 & 2.0\,min \\
GeDi, whole scan & 350,000 & 5\,cm & 33.3 & 36.7 & 2.0\,min \\
GeDi, whole scan & 350,000 & 10\,cm & 40.0 & 43.3 & 2.1\,min \\
GeDi, whole scan & 350,000 & 15\,cm & 43.3 & 46.7 & 2.0\,min \\
GeDi, whole scan & 350,000 & 20\,cm & 40.0 & 43.3 & 2.1\,min \\
GeDi, whole scan & 1,000,000 & 5\,cm & 36.7 & 43.3 & 5.4\,min \\
GeDi, whole scan & 1,000,000 & 10\,cm & 40.0 & 43.3 & 5.2\,min \\
GeDi, whole scan & every point & 2\,cm$^{\ast}$ & 26.7 & 30.0 & 8.9\,min \\
\midrule
GeDi, 1,090 cubes & every point & 2\,cm$^{\ast}$ & 76.7 & 86.7 & 91\,min \\
BUFFER-X, 1,090 cubes & every point & released & 60.0 & 63.3 & 16\,min \\
\oursrow \ours \fast, whole scan & 350,087 cells & -- & 84.4 & 90.0 &\oursend \best{0.38}\,s \\
\oursrow \ours \precise, whole scan & 350,087 cells & -- & \best{86.7} & \best{91.1} &\oursend 0.62\,s \\
\bottomrule
\end{tabular}

\caption{\textbf{The tested GeDi settings on the whole scan remain below its cube search and \ours.}
All rows use the first frame of each walk, \pnum{hosp.sub_frames} rendered frames with simulated depth noise and matching colors.
Baselines include the shared refinement.
Recall (\%) uses 5\,cm and 5\textdegree{} or 0.5\,m and 5\textdegree.
Gray rows are \ours, and bold marks the best value per column.
The keypoint column gives how many points GeDi describes, or the cell count for \ours.
The inlier distance sets the match tolerance, and $^\ast$ marks the room-calibrated value.
Time is the median per frame on an RTX 2080 Ti.}
\label{tab:hospital_grid}
\end{table}

\begin{table}[t]
\centering
\scriptsize\setlength{\tabcolsep}{2.5pt}
\begin{tabular}{@{}lrrrrr@{}}
\toprule
Region searched & Map points & Frame inside & 5\,cm & 0.25\,m & 0.5\,m \\
\midrule
3\,m cube around the seen points & 24,570 & 75\% & 50.0 & 66.7 & \best{83.3} \\
6\,m cube around the seen points & 103,354 & 99\% & 30.0 & 50.0 & 66.7 \\
4\,m square around the camera & 53,413 & 39\% & 23.3 & 30.0 & 40.0 \\
8\,m square around the camera & 174,401 & 89\% & 30.0 & 36.7 & 53.3 \\
16\,m square around the camera & 628,505 & 100\% & 23.3 & 23.3 & 36.7 \\
32\,m square around the camera & 1,537,817 & 100\% & 13.3 & 20.0 & 26.7 \\
\midrule
Whole scan & 1,917,644 & 100\% & 20.0 & 13.3 & 26.7 \\
1,090 cubes of 3\,m, best cube & every point & -- & \best{63.3} & \best{80.0} & \best{83.3} \\
\bottomrule
\end{tabular}

\caption{\textbf{GeDi's recall falls steadily as the searched region grows.}
GeDi uses its calibrated settings and no refinement, on the first frame of each walk.
Regions are cut from the scan with the true pose, so the rows diagnose the search and are not a localization result.
Cubes center on the mean of the frame's points, and squares center on the camera and span every height.
Frame inside gives the median share of the frame's points inside the region, which limits the smallest square.
The cube search keeps, per frame, the cube with the most agreeing matches.
Recall (\%) uses 5\,cm and 5\textdegree, 0.25\,m and 2\textdegree, or 0.5\,m and 5\textdegree.
Bold marks the best value per column.}
\label{tab:hospital_crops}
\end{table}

\section{BUFFER-X and the size of the point cloud}
\label{sec:supp_bufferx}

BUFFER-X~\cite{seo2025bufferx} registers two partially overlapping point clouds.
It derives its patch radii from the larger input point cloud. Its released code caps them at \pnum{bx.cap_m}\,m and samples a fixed number of keypoints from each input.
As the reference point cloud grows, the patches grow and the keypoint budget spreads across more space.
On the grid of rooms in \cref{sec:supp_scale}, the median largest radius rises from \pnum{bx.k1.r_big}\,m for one room to \pnum{bx.k2.r_big}\,m for two and \pnum{bx.k5.r_big}\,m for five.
It reaches the cap from \pnum{bx.cap_from_rooms} rooms onward.
At this scale, patches describe rooms rather than local surfaces.
The paper therefore evaluates BUFFER-X only on single rooms and \pnum{hosp.tile_m}\,m cubes.

Taking the radii from the frame instead does not improve the room-given result.
On \pnum{bx.scale.frames} ScanNet-10 frames, this reduces the median largest radius from \pnum{bx.scale.official.r_big} to \pnum{bx.scale.frame.r_big}\,m.
With shared refinement, recall falls from \pnum{bx.scale.official.r5}\% to \pnum{bx.scale.frame.r5}\% at 5\,cm and 5\textdegree, and from \pnum{bx.scale.official.r50}\% to \pnum{bx.scale.frame.r50}\% at 0.5\,m and 5\textdegree.
The room-given comparisons retain the released radius rule.

\section{A point cloud colored by another camera}
\label{sec:supp_bonn}

In Bonn~\cite{palazzolo2019refusion}, the query frames and the point cloud come from different sensors.
Its \pnum{bonn.frames} test frames come from the static ASUS Xtion RGB-D sequence, with motion-capture reference poses.
The same room is represented by a terrestrial laser scan colored by the scanner's camera.
\ours searches the full point cloud with its room-unknown settings.
GeDi searches both the full point cloud and overlapping \pnum{hosp.tile_m}\,m cubes.
BUFFER-X receives only cubes, following \cref{sec:supp_bufferx}.

The cameras record the same walls in different colors, and \ours has low recall (\cref{tab:bonn}).
\fast reaches \pnum{bonn.fast.r50}\% within 0.5\,m and 5\textdegree, or \pnum{bonn.fast_color.r50}\% with the \colorpass.
GeDi, which does not use color, reaches \pnum{bonn.gedi_whole.r50}\% on the whole point cloud and \pnum{bonn.gedi_tiles.r50}\% on cubes.
Its median times are \pnum{bonn.gedi_whole.min} and \pnum{bonn.gedi_tiles.min} minutes per frame respectively.
This is the boundary of the main paper's gains: ScanNet, 7-Scenes and 12-Scenes use the same kind of camera for frames and point clouds, while the hospital's rendered colors agree by construction.

\begin{table}[t]
\centering
\scriptsize\setlength{\tabcolsep}{3pt}
\begin{tabular}{@{}lrrrr@{}}
\toprule
Method & 5\,cm & 0.25\,m & 0.5\,m & Time \\
\midrule
\oursrow \ours \fast & 9.1 & 14.5 & 26.3 &\oursend \\
\oursrow \ours \precise & 10.4 & 14.5 & 27.2 &\oursend \\
\oursrow \ours \fast + \colorpass & 17.3 & 27.6 & 47.2 &\oursend \\
\oursrow \ours \precise + \colorpass & 18.3 & 27.5 & 48.6 &\oursend \\
\midrule
GeDi~\cite{poiesi2022learning}, whole scan & 36.7 & 50.9 & 77.1 & 2.8\,min \\
GeDi, 3\,m cubes & \best{42.2} & \best{61.3} & \best{91.5} & 11.0\,min \\
BUFFER-X~\cite{seo2025bufferx}, 3\,m cubes & 34.5 & 49.1 & 74.5 & \best{1.4}\,min \\
\bottomrule
\end{tabular}

\caption{\textbf{Point cloud registration methods retain the lead across cameras.}
Recall (\%) on Bonn uses 5\,cm and 5\textdegree, 0.25\,m and 2\textdegree, or 0.5\,m and 5\textdegree.
Gray rows are \ours, and bold marks the best value per column.
\ours averages three training seeds.
Baselines include the shared refinement.
Baseline times are medians per frame on an RTX 2080 Ti.}
\label{tab:bonn}
\end{table}

\begin{table*}[t]
\centering
\scriptsize\setlength{\tabcolsep}{3pt}
\begin{tabular}{@{}l*{13}{r}@{}}
\toprule
 & & \multicolumn{2}{c}{Narrower lens} & \multicolumn{4}{c}{Focal length error} & \multicolumn{4}{c}{Radial distortion $k_1$} & \multicolumn{2}{c@{}}{Color offset} \\
\cmidrule(lr){3-4}\cmidrule(lr){5-8}\cmidrule(lr){9-12}\cmidrule(l){13-14}
Method & Unchanged & 45\textdegree & 30\textdegree & +5\% & +10\% & $-$5\% & $-$10\% & +0.1 & +0.2 & $-$0.1 & $-$0.2 & 5\,px & 10\,px \\
\midrule
\multicolumn{14}{@{}l}{\emph{Recall at 5\,cm and 5\textdegree}} \\
\oursrow \ours \fast & 75.6 & \best{71.8} & \best{60.9} & 56.4 & 21.7 & \best{53.9} & \best{16.5} & 73.1 & \best{63.9} & 73.0 & \best{62.5} & 74.9 &\oursend 75.4 \\
\oursrow \ours \precise & 79.1 & 75.3 & 63.9 & 51.2 & 16.3 & 47.5 & 11.7 & 76.9 & 67.4 & 75.0 & 61.8 & 78.6 &\oursend 78.5 \\
BUFFER-X~\cite{seo2025bufferx} & 74.4 & 69.7 & 52.0 & 56.3 & 21.5 & 53.6 & 16.2 & 72.0 & 62.9 & 71.4 & 61.7 & 74.4 & 74.4 \\
GeDi~\cite{poiesi2022learning} & \best{76.2} & 71.6 & 54.9 & \best{56.8} & \best{21.9} & 53.6 & \best{16.5} & \best{73.2} & 63.8 & \best{73.4} & 62.3 & \best{76.2} & \best{76.2} \\
\midrule
\multicolumn{14}{@{}l}{\emph{Recall at 0.25\,m and 2\textdegree}} \\
\oursrow \ours \fast & 86.9 & \best{80.9} & \best{68.6} & 75.1 & \best{30.3} & 74.4 & \best{17.9} & 85.5 & 79.4 & 85.8 & 78.9 & 86.5 &\oursend 86.5 \\
\oursrow \ours \precise & 89.1 & 83.0 & 71.0 & 74.1 & 23.3 & 68.8 & 12.1 & 86.5 & 81.3 & 85.9 & 77.7 & 88.7 &\oursend 88.5 \\
BUFFER-X~\cite{seo2025bufferx} & 86.2 & 78.6 & 59.0 & 74.6 & \best{30.3} & \best{74.9} & 17.2 & 84.7 & 77.6 & 85.3 & 77.6 & 86.2 & 86.2 \\
GeDi~\cite{poiesi2022learning} & \best{87.3} & 80.7 & 62.5 & \best{76.0} & 29.5 & 74.4 & 17.2 & \best{85.6} & \best{79.7} & \best{86.8} & \best{79.7} & \best{87.3} & \best{87.3} \\
\bottomrule
\end{tabular}

\caption{\textbf{Calibration errors cost every method alike, and a narrow lens favors \ours.}
Recall (\%) on the \pnum{cam.frames} real ScanNet-10 frames when one property of the query camera changes.
A focal length error of +\pnum{cam.focal.small}\% means a focal length \pnum{cam.focal.small}\% too long.
Gray rows are \ours.
Bold marks the best value per column within each block among \fast and the baselines, which share one refinement.
\precise adds a second refinement stage.
\ours averages three training seeds.
A color offset leaves the geometry unchanged, so the baselines repeat their unchanged values there.}
\label{tab:camera}
\end{table*}

\section{Camera and lens properties}
\label{sec:supp_camera}

A deployed camera can differ from the development cameras in its lens and its calibration.
We change one property of the query camera at a time on the \pnum{cam.frames} real ScanNet-10 frames and leave the colored point cloud unchanged (\cref{tab:camera}).
A narrower lens keeps only the points inside a horizontal field of view of \pnum{cam.fov45.deg}\textdegree{} or \pnum{cam.fov30.deg}\textdegree, where the ScanNet depth camera sees \pnum{cam.native_hfov}\textdegree.
A focal length \pnum{cam.focal.small} or \pnum{cam.focal.large}\% too long or too short scales each frame across the optical axis.
Uncorrected radial distortion with coefficient $k_1$ moves each point away from the image center for positive $k_1$ (pincushion) or toward it for negative $k_1$ (barrel), in proportion to its squared distance from the center.
A color offset pairs each depth pixel with the color \pnum{cam.color.small} or \pnum{cam.color.large} pixels to its right.
BUFFER-X and GeDi receive the same frames and the shared refinement.

Calibration errors distort the frame itself, so they cost every method alike.
With a focal length \pnum{cam.focal.large}\% too long, \fast localizes \pnum{cam.fast.focal110.r25}\% of the frames within 0.25\,m and 2\textdegree, GeDi \pnum{cam.gedi.focal110.r25}\% and BUFFER-X \pnum{cam.bx.focal110.r25}\%, against \pnum{cam.fast.clean.r25}\% for \fast on unchanged frames.
The place is still found: within 0.5\,m and 5\textdegree, \fast keeps at least \pnum{cam.fast.focal.min_r50}\% at every focal error.
The scaled frame cannot fit the point cloud closely, and the second refinement stage of \precise ends below \fast at all \pnum{cam.precise_below_fast.focal} focal errors.
Distortion costs \fast at most \pnum{cam.fast.dist.max_drop} points.
In \pnum{cam.calib.equivalent} of \pnum{cam.calib.comparisons} comparisons over the focal and distortion levels, the losses of \fast and of a baseline are equivalent within \pnum{cam.margin} points (Sec.~\ref{sec:setup}).
A color offset of up to \pnum{cam.color.large} pixels changes \fast by at most \pnum{cam.fast.color.max_change} points.

A narrow lens separates the methods.
A \pnum{cam.fov30.deg}\textdegree{} lens keeps \pnum{cam.fov30.kept}\% of the frame's points.
\fast then leads GeDi by \pnum{cam.fov30.fast_minus_gedi} points [\pnum{cam.fov30.fast_minus_gedi.lo}, \pnum{cam.fov30.fast_minus_gedi.hi}] and BUFFER-X by \pnum{cam.fov30.fast_minus_bx} [\pnum{cam.fov30.fast_minus_bx.lo}, \pnum{cam.fov30.fast_minus_bx.hi}].

\end{document}